\documentclass[journal]{IEEEtran}
\usepackage{algorithmic}
\usepackage{algorithm}
\usepackage{array}
\usepackage{textcomp}
\usepackage{stfloats}
\usepackage{url}
\usepackage{verbatim}
\usepackage{graphicx}
\usepackage{cite}
\usepackage{amsmath,amssymb,amsfonts}
\usepackage{xcolor}
\usepackage[caption=false]{subfig}
\usepackage{multirow}
\usepackage{tabularx}
\usepackage{booktabs}
\usepackage{comment}
\usepackage{bm}
\usepackage{makecell}
\def\BibTeX{{\rm B\kern-.05em{\sc i\kern-.025em b}\kern-.08em
    T\kern-.1667em\lower.7ex\hbox{E}\kern-.125emX}}

\newcolumntype{Y}{>{\centering\arraybackslash}X}

\begin{document}

\title{Enabling Federated Inference via Unsupervised Consensus Embedding}

\author{Yui Hashimoto, Takayuki Nishio,~\IEEEmembership{Senior Member,~IEEE,}
Yuichi Kitagawa,
and Takahito Tanimura
\thanks{Yui Hashimoto and Takayuki Nishio are with the School of Engineering, Institute of Science Tokyo, Tokyo 152-8550, Japan (e-mail: nishio@
ict.eng.isct.ac.jp). Yuichi Kitagawa and Takahito Tanimura are with the
Research \& Development Group, Hitachi Ltd., Tokyo 185-8601, Japan. 

This work has been submitted to the IEEE for possible publication. Copyright may be transferred without notice, after which this version may no longer be accessible.}
}

\maketitle

\begin{abstract}
Cooperative inference across independently deployed machine learning models is increasingly desirable in distributed environments, as there is a growing need to leverage multiple models while keeping their data and model parameters private. However, existing cooperative frameworks typically rely on sharing input data, model parameters, or a common encoder, which limits their applicability in privacy-sensitive or cross-organizational settings.
To address this challenge, we propose \textbf{Consensus Embedding-based Federated Inference (CE-FI)}, a framework that enables pretrained models to cooperate at inference time without sharing model parameters or raw inputs and without assuming a common encoder. CE-FI introduces two components: a Consensus Embedding (CE) layer that maps heterogeneous intermediate representations into a common embedding space, and a Cooperative Output (CO) layer that produces predictions from these embeddings. Both layers are trained using shared unlabeled data only, so the cooperative stage does not require additional labeled data.
Experiments on image classification benchmarks—CIFAR-10 and CIFAR-100—under diverse non-IID conditions show that CE-FI consistently outperforms solo inference and performs comparably to conventional methods that require stronger sharing assumptions. Additional evaluations on text and time-series tasks indicate applicability beyond image classification, although performance depends on the ensemble strategy. Further analysis identifies representation alignment as the primary bottleneck.
\end{abstract}


\section{Introduction}
With the widespread deployment of AI technologies, high-performance inference models are now readily available across diverse environments. Under such circumstances, there is a growing demand for frameworks that can cooperatively leverage multiple existing models to improve inference performance. This concept of model collaboration has become increasingly important across a wide range of domains, including large language models, reflecting a broader trend in modern AI systems toward reusing pretrained models to achieve strong inference capabilities~\cite{modelmerging-akiba2025evolutionary,moe-6797059}.
At the same time, models deployed in different environments are typically designed and operated independently, and their parameters are often treated as intellectual property or confidential assets. As a result, simply aggregating or distributing pretrained models is frequently infeasible due to licensing constraints and organizational governance. Moreover, recent studies have reported that sharing model parameters may enable the reconstruction or inference of training data (e.g., through model inversion attacks~\cite{fredrik2015inversion}). These concerns motivate the need for cooperative inference frameworks that do not require model sharing.

\begin{figure}[t!]
\centering
\includegraphics[width=\linewidth]{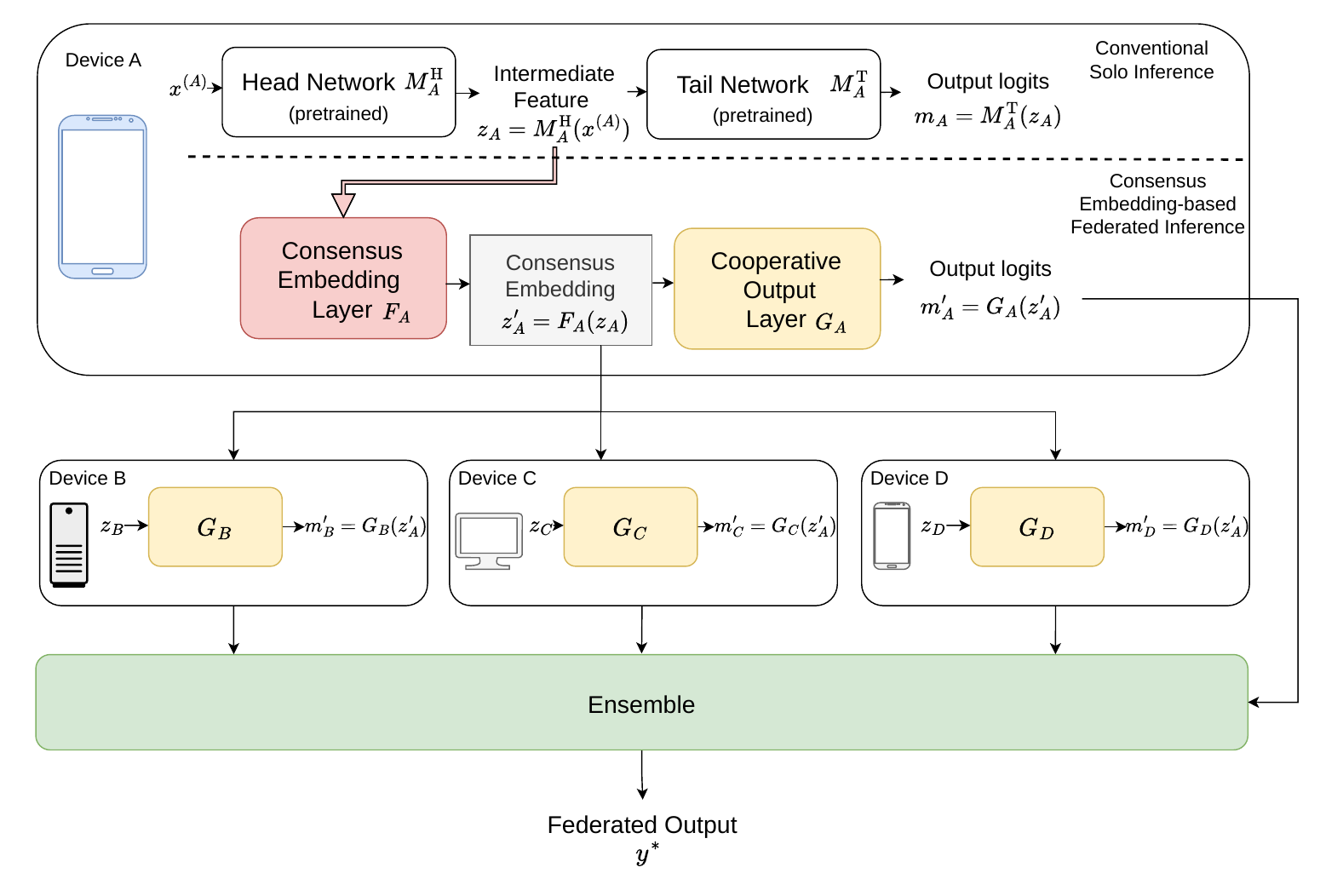}
\caption{Cooperative inference with CE-FI. Each device maps the intermediate feature of an input sample into a shared feature space via the Consensus Embedding Layer, shares the resulting consensus embeddings with other cooperative devices, performs inference based on the received embeddings, and integrates all outputs to obtain improved predictions.}
\label{fig:framework_fig}
\end{figure}

In this context, Federated Inference (FI)~\cite{fi-zhou2025towards} has been established as a paradigm in which clients retain their pretrained models locally while collaborating only at the inference stage. In FI, an inference request—the information required for prediction—is shared with multiple participating clients, each of which performs inference using its local model and returns prediction results for aggregation. This design allows FI to enable cooperation without exposing model parameters and also decouples cooperative inference from training, making it possible to reuse existing models without retraining.
In a representative FI setting, all participants receive the same input, and each model independently produces predictions based on the shared input. While this is one of the most fundamental forms of FI and is widely considered in prior works~\cite{weng2021fedserv,coopinf-11165313}, sharing raw input data is often impractical under strict privacy and confidentiality constraints. For example, in medical applications, sensitive information such as patient records or diagnostic images cannot be directly shared due to legal and ethical restrictions. In such environments, FI configurations that require input sharing cannot be directly applied.

Edge Ensemble~\cite{malka2025edgeensemble} represents a different FI configuration in which cooperation is enabled through intermediate representations produced by a commonly adopted encoder architecture. While this approach avoids distributing raw inputs explicitly, it still relies on the availability of a shared representation space across devices. 
In practice, pretrained models are often developed under different architectures and training procedures, resulting in heterogeneous feature spaces whose intermediate representations are not directly compatible~\cite{svcaa-nips2017,li2015convergent}. Furthermore, deploying a shared encoder would require disclosing part of a model's parameters, which is not acceptable in environments where even encoder sharing is prohibited.

Therefore, while FI provides an effective means of cooperative  inference without model sharing, existing approaches still rely on the existence of a common input or representation space. When such assumptions do not hold, realizing cooperative inference becomes highly challenging.

Motivated by this limitation, this work addresses the following question:
\textit{How can FI-style cooperative  inference be achieved across independently trained models without sharing inputs and without assuming a shared encoder?
}
This setting breaks the standard FI assumption that multiple clients can produce predictions for the same input.

To tackle this challenge, we propose Consensus Embedding-based Federated Inference (CE-FI). In heterogeneous environments, each pretrained model forms its own internal representation space, making intermediate features difficult to reuse directly across devices. To address this issue, the key idea of CE-FI is to learn a common feature space for cooperation by mapping each model's intermediate representations into a device-shareable consensus embedding space. Representations in this shared space, referred to as consensus embeddings, provide a minimal interface for knowledge exchange across heterogeneous models without requiring raw input or parameter sharing.
As illustrated in Fig.~\ref{fig:framework_fig}, CE-FI consists of (i) a Consensus Embedding (CE) layer, which aligns heterogeneous intermediate features into a shared space, and (ii) a Cooperative Output (CO) layer, which reconstructs device-specific predictions from the consensus embeddings for final inference aggregation. Through these components, CE-FI preserves the core structure of FI while eliminating the need for input sharing or common encoders.
Furthermore, CE-FI does not require labeled data for additional cooperative training. Both the CE and CO layers are trained using unlabeled data that is commonly accessible across devices. Specifically, the CE layer is optimized via label-free contrastive learning to align embeddings, while the CO layer is trained through unlabeled knowledge distillation using soft targets produced by the local pretrained models. 

We evaluate CE-FI on image classification tasks using pretrained encoders on CIFAR-10 and CIFAR-100 under various non-IID conditions. Results show that CE-FI consistently outperforms solo inference and, in some cases, approaches the performance of conventional FI methods that permit input sharing or common encoders. We further extend the evaluation to other modalities, including text classification on IMDB and time-series recognition on UCI-HAR, to examine the applicability of CE-FI. Ablation studies reveal that output aggregation strategies critically impact performance and that embedding alignment is a major bottleneck. Additionally, reconstruction experiments suggest that sharing intermediate features and consensus embeddings does not trivially reveal raw inputs under the examined settings.

The main contributions of this paper are summarized as follows:

\vspace{1mm}
\noindent\textbf{Cooperative  inference without inputs or model sharing:} 
Unlike existing FI approaches, CE-FI achieves cooperative inference without sharing model parameters or raw inputs, and without assuming a common encoder. As a result, CE-FI naturally supports cooperation across heterogeneous pretrained models. This is accomplished solely through intermediate feature exchange and self-supervised learning, allowing CE-FI to address these constraints simultaneously.

\vspace{1mm}
\noindent\textbf{Self-supervised training of embedding and prediction layers:} 
CE-FI introduces a contrastive-learning-based embedding layer and a knowledge-distillation-based prediction layer, both trained with unlabeled data. This allows the system to operate in label-scarce environments, reducing annotation costs and enhancing deployment flexibility.

\vspace{1mm}
\noindent\textbf{Theoretical characterization of FI consistency:}
We formally analyze the relationship between CE-FI and conventional FI, showing that CE-FI retains structural consistency with FI and can theoretically preserve cooperative inference performance even under stricter constraints.

\vspace{1mm}
\noindent\textbf{Comprehensive evaluation across distributed settings and modalities:} 
We conduct extensive experiments across multiple tasks, modalities, and non-IID conditions under a cross-silo scenario with 3--10 devices. The results demonstrate that CE-FI consistently improves over solo inference and can approach conventional FI performance, highlighting its effectiveness under diverse settings for privacy-sensitive distributed AI.

\section{Related Work}

\begin{table*}[t]
\centering
\caption{Comparison between CE-FI and existing cooperative learning/inference frameworks. \checkmark\ indicates a desirable property.}
\label{tab:comparison}
\begin{tabular}{lccccccc}
\hline
\makecell{\textbf{Method}} & 
\makecell{\textbf{Framework}} & \makecell{\textbf{Pre-trained}\\ \textbf{Models}} & 
\makecell{\textbf{Multi-user}\\ \textbf{Coop.}} & 
\makecell{\textbf{No Model}\\ \textbf{Sharing}} & 
\makecell{\textbf{No Input}\\ \textbf{Sharing}} &
\makecell{\textbf{No Shared}\\ \textbf{Encoder Req.}} & 
\makecell{\textbf{No Labeled}\\ \textbf{Dataset Req.}} \\[1.5pt]
\hline
Federated Learning& Learning & $\triangle$ & \checkmark &  & \checkmark & \checkmark \\
Distillation-based FL & Learning & $\triangle$ & \checkmark & \checkmark & \checkmark & \checkmark &  \\
Transfer Learning & Learning & \checkmark & $\triangle$  &  & \checkmark & \checkmark \\
One-shot FL & Learning & \checkmark & \checkmark  &  &  \checkmark  & \checkmark & $\triangle$\\
Input-sharing FI & Inference & \checkmark & \checkmark & \checkmark  &  &\checkmark & \checkmark\\
Edge Ensemble & Inference & \checkmark & \checkmark &  \checkmark  &\checkmark & & \checkmark \\
\textbf{CE-FI (Ours)} & Inference & \checkmark & \checkmark & \checkmark & \checkmark & \checkmark & \checkmark \\
\hline
\end{tabular}
\end{table*}

The idea of improving AI model performance by cooperatively leveraging multiple models has been extensively studied. One of the earliest and most widely used approaches is ensemble learning~\cite{hansen1990neural}, which combines multiple models trained under different conditions for the same task, thereby improving accuracy and generalization ability by integrating their outputs. By combining diverse models, ensemble learning mitigates individual errors and improves overall performance. Techniques such as bagging and boosting further advance this principle by resampling training data and adjusting model weights to achieve both diversity and accuracy~\cite{dietterich2000ensemble}. However, these methods assume centralized management of models and data, making them difficult to apply in distributed environments with privacy or communication constraints.

To relax the assumption of centralized data collection, Federated Learning (FL) has been proposed~\cite{mcmahan2017communication}.  In FL, each client trains a model locally using private data, and a central server aggregates the resulting local updates to achieve cooperative learning while preserving data confidentiality. This framework represents a major advance in enabling distributed learning and has been widely studied as a foundation of privacy-aware machine learning~\cite{yang2019federated}.
Subsequent work has extended FL to address practical challenges such as statistical heterogeneity and communication efficiency, including FedProx~\cite{li2020fedprox} and MOON~\cite{li2021moon}. 
These methods improve robustness under heterogeneous data distributions while preserving the federated training paradigm. 
Nonetheless, FL generally requires all clients to adopt the same model architecture, making it unsuitable for environments with heterogeneous models. Furthermore, clients must share locally updated models with at least the server—or with other clients in some algorithms—which introduces the risk of model inversion. Therefore, in scenarios where model parameters cannot be disclosed, FL-based cooperation at the training stage becomes fundamentally constrained.

Knowledge distillation~\cite{hinton2015distilling} has been widely explored as a mechanism for cooperative learning across heterogeneous architectures. In this method, a high-performing teacher model provides soft targets used to train a student model, facilitating knowledge transfer even across heterogeneous architectures. This allows lightweight models to benefit from the capabilities of more complex ones. Furthermore, Co-distillation~\cite{anil2018large} allows multiple student models to learn from each other's outputs without relying on a specific teacher model, enabling mutual knowledge sharing. Additionally, methods such as FedKD~\cite{wu2022communication} and DS-FL~\cite{itahara2023DSFL} integrate distillation into federated learning, improving communication efficiency and enabling cooperation across heterogeneous models.
However, these methods often depend on labeled data and may still require sharing model-related information during training. As a result, they remain difficult to deploy in environments where model disclosure is prohibited and only unlabeled data are available.

Another line of work reuses pretrained models without fully retraining them. Transfer Learning~\cite{5288526,sharif2014cnn-transfer} aims to reuse the feature extraction layers of a model trained on a source task for a different target task, often with fine-tuning of an added output layer. This approach improves learning efficiency and supports adaptation in low-data regimes.  However, this method requires access to the source model's parameters or features, which may not be permissible in certain settings.
Additionally, One-Shot Federated Learning~\cite{guha2019one,AMATO2026one-shot-survey} has been proposed to reduce the communication costs inherent in FL. In this approach, each client sends its locally trained model to the central server only once, after which ensemble learning or knowledge distillation is used to construct a global model. While this reduces communication costs compared to conventional FL, it still requires central aggregation of model parameters, which is unsuitable for applications involving highly confidential models.

In contrast, Federated Inference (FI) has emerged as a framework for achieving cooperative inference while retaining pretrained models locally, without requiring model sharing or retraining. In FI, each participating device performs inference using its own pretrained model, and their prediction results are aggregated to enable cooperation. This paradigm avoids centralized model collection and preserves model confidentiality, making it highly relevant for practical deployment. Zhou \textit{et al.}~\cite{fi-zhou2025towards} systematically formalized FI while addressing the optimization problem of participant selection, clarifying its conceptual definition and fundamental structure. One common FI configuration assumes that a central node shares the input data, allowing each model to independently generate predictions for the same input, which are then aggregated. Many FI-related studies adopt this setting, such as  FedServing~\cite{weng2021fedserv}, which investigates system-level design and operational challenges for cooperative inference services. 

However, sharing raw input data is often infeasible in privacy-sensitive environments. To relax this requirement, FI can also operate over a shared representation space. Edge Ensemble~\cite{malka2025edgeensemble} exemplifies this encoder-shared  configuration, where cooperation is achieved through intermediate features extracted by a common encoder and quantizer. In fact, this setting subsumes input-sharing FI as a special case when both the encoder and quantizer  reduce to an identity mapping. 
Nevertheless, this approach critically assumes that all participating models share the same encoder and thus operate in a common representation space. In practice, pretrained models often possess heterogeneous feature spaces, and in many scenarios, even partial disclosure of encoder parameters may be undesirable under strict governance or confidentiality constraints. Consequently, Edge Ensemble cannot be applied when models are heterogeneous or when any model sharing is prohibited. Overall, while existing FI approaches consistently avoid sharing full model parameters, they still rely on either input sharing or a common encoder. Under strict constraints where neither raw inputs can be shared nor a common encoder can be assumed, the applicability of conventional FI remains limited.

To clarify the position of CE-FI among existing methods, Table~\ref{tab:comparison} summarizes the key differences in terms of required resources and applicable settings.
As illustrated above, most existing cooperative learning and cooperative inference frameworks assume that some form of data, model, label information, or a shared encoder is available at either the training or inference stage. While such assumptions hold in many practical scenarios, they may break down due to privacy, regulatory, or organizational constraints that limit the availability or exchange of these resources.
In contrast, this work targets a significantly stricter setting, where neither a shared input space nor model parameters can be exchanged. Under these constraints, the shared foundations of conventional approaches must be fundamentally reconsidered. The proposed CE-FI framework addresses this challenge by leveraging only intermediate features from local models and unlabeled shared data, without requiring input sharing or model disclosure. 
This design enables flexible cooperative inference across heterogeneous devices while maintaining strong inference performance.
Therefore, CE-FI removes the input-sharing and shared-encoder assumptions of conventional FI and extends FI toward a new cooperative inference paradigm under stricter sharing constraints. This provides a promising framework for practical deployment in distributed intelligent systems.
\section{Consensus Embedding-based Federated Inference}
\subsection{System Model}
We consider a cooperative distributed computing environment in which edge devices—such as smartphones, laptops, and local servers—interact over a network to perform inference. Each device $k$ maintains a local pretrained model $M_k$, which consists of a head network $M^\mathrm{H}_k$ that maps an input sample $x$ into an intermediate feature representation, and a tail network $M^\mathrm{T}_k$ that produces task-specific predictions based on this representation. The model $M_k$ is pretrained using the device’s private dataset $D_k$, which is used only for the initial training stage.
We assume that each pretrained model $M_k$ exhibits different classification capabilities and class coverage, reflecting the label distribution of its local dataset $D_k$. For example, in a 10-class classification task, device 1 may achieve high accuracy for classes 0–6, device 2 for classes 3–9, and device 3 for classes 0–2 and 6–9. 

We assume that the head network $M^\mathrm{H}_k$ is implemented as a task-dependent pretrained encoder. For instance, in image classification, large-scale pretrained models such as Vision Transformer (ViT)~\cite{vit-dosovitskiy2021an} can serve as the head network. When such pretrained encoders are unavailable, the head network may alternatively be defined as a partial subnetwork of a model trained from scratch on the private dataset $D_k$, corresponding to the layers up to a certain intermediate depth.

From a privacy and confidentiality standpoint, we assume that devices cannot share the architecture or parameters of model $M_k$. This assumption reflects practical constraints in many applications—such as in the medical and financial domains—where trained models may contain proprietary knowledge or statistical characteristics derived from sensitive training data. Consequently, organizations may be unwilling or unable to disclose model architectures or parameters due to privacy, regulatory, or intellectual property concerns. 
Instead, we assume that intermediate features from each model can be shared with other devices. 
Specifically, although raw input data $x$ cannot be shared, the intermediate representation  ${z_{k,x}}=M^\mathrm{H}_k(x)$ obtained by processing $x$ through the head network of the pretrained model can be shared. This assumption is widely adopted in prior work on Split Learning and Split Computing~\cite{gupta2018distributed,kang2017neurosurgeon}, where intermediate features are considered less directly identifiable than raw inputs.

Furthermore, during the training phase of CE-FI, we assume the existence of a shared unlabeled dataset $D_\mathrm{share}$ that is accessible to all devices. The use of such shared datasets is commonly adopted
in prior studies, including distillation-based FL frameworks~\cite{itahara2023DSFL}. The availability of such shared unlabeled data is also a reasonable assumption in real-world edge-cloud cooperative systems, where non-personal unlabeled data, such as log data or sensor data, is often shared via the cloud. In these configurations, data collection in forms that exclude sensitive information is generally permissible under institutional regulations, making this assumption practically valid.

\subsection{Overview of Proposed Method}
We propose a novel cooperative inference method, Consensus Embedding-based Federated Inference (CE-FI), which enables multiple devices with pretrained models to cooperatively perform inference and improve prediction accuracy in distributed environments—without input sharing, model sharing, or a common encoder. CE-FI also introduces a self-supervised training strategy for the required layers, eliminating the need for any labeled dataset.

The cooperative inference process of CE-FI is illustrated in Fig.~\ref{fig:framework_fig}.
Given an input sample $x^{(k)}$ at device $k$, which cannot be shared with other devices, the head network of device $k$ produces an intermediate feature ${z_{k,x}} = M^\mathrm{H}_k(x^{(k)})$. 
In standard inference, this intermediate feature $z_{k,x}$ is passed to the device's own tail network, which then outputs the logits $m_{k,x} = M^\mathrm{T}_k(z_{k,x})$. However, directly passing $z_{k,x}$ to other devices and inputting it into their tail networks does not lead to correct predictions, since $z_{k,x}$ belongs to a device-specific feature space. Even for the same input, we typically observe $\lVert z_{i,x} - z_{j,x}\lVert \gg 0$ for $i \neq j$. 

To address this challenge, CE-FI introduces a Consensus Embedding (CE) layer $F_k(\cdot)$ on each device, which transforms the intermediate feature $z_{k,x}$ into a common representation $z_{k,x}' = F_k(z_{k,x})$. This transformed feature, referred to as the consensus embedding, resides in a common embedding space and is designed to be semantically aligned across devices, allowing it to be processed by any device.
Each device is therefore also equipped with a Cooperative Output (CO) layer $G_k(\cdot)$, which maps the consensus embedding to output logits $m_{k,x}' = G_k(z_{k,x}')$. For notational simplicity, we may omit the subscript $x$ in the following discussion (e.g., $z_{k,x}$ is denoted as $z_k$ and $m_{k,x}$ as $m_k$). The training procedures for both the CE and CO layers are designed to be self-supervised and do not require any labeled data; details are provided in the following subsection.

To obtain the final prediction, the output logits $m_k'$ from each device are aggregated through an ensemble mechanism. Several strategies can be considered, including hard voting and soft voting, which are commonly used in conventional ensemble learning~\cite{gandhi2015hard,sherazi2021soft}. In this study, we adopt a confidence-based selection approach, where a confidence score is estimated for each output $m_k'$, and the final prediction $y^*$ is determined by selecting the output with the highest confidence. The details of the ensemble strategy are presented in Sect.~\ref{subsec:inference}.
Through this mechanism, CE-FI enables cooperative inference across multiple devices using consensus embeddings.

\subsection{Unsupervised Training of Consensus Embedding Layer}

\begin{figure}
    \centering
    \includegraphics[width=\linewidth]{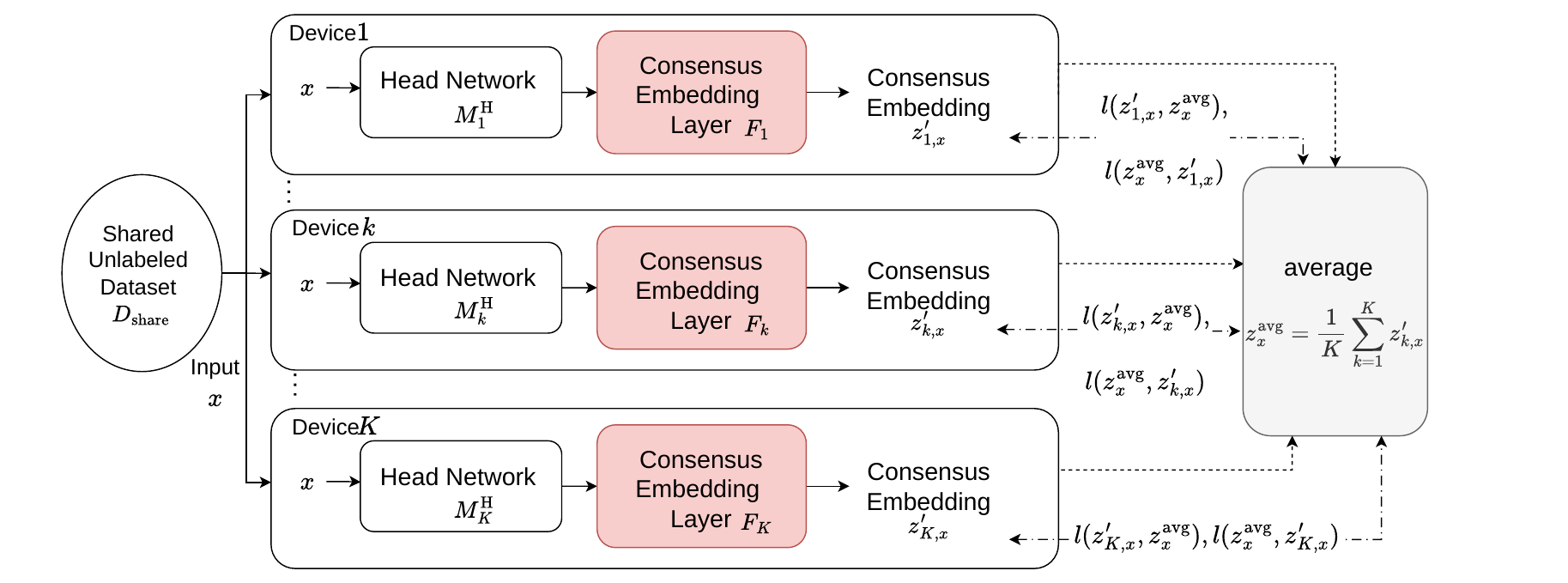}
    \caption{Overview of the CE layer training procedure. Each device generates a consensus embedding for the same input sample; these embeddings are aggregated, and a contrastive loss is computed to align them across devices.}
    \label{fig:study-method-consensus}
\end{figure}

Fig.~\ref{fig:study-method-consensus} illustrates the overall training process for the CE layer. The primary goal is to align the consensus embeddings generated by different devices so that they converge to a common representation for the same input. During training, each device retains its own pretrained model, and the only data available is the unlabeled dataset $D_\mathrm{share}$, which is shared across all devices. The parameters $\theta_{F_k}$ of the CE layer $F_k$ are optimized by minimizing a contrastive loss $L_\mathrm{cons}$, which requires no label information and encourages semantic consistency by pulling together consensus embeddings derived from the same input and pushing apart those from different inputs.

Specifically, each device $k$ first computes an intermediate feature ${z_{k,x}} = M^\mathrm{H}_k(x)$ by passing each sample $x \in D_\mathrm{share}$ through the head network of its pretrained model. These features are then fed into the CE layer to obtain the corresponding consensus embeddings $z'_{k,x} = F_k({z_{k,x}})$. The contrastive loss $L_\mathrm{cons}$ is computed based on these embeddings (its formulation is described later), and the resulting gradient $\nabla_{\theta_{F_k}}L_\mathrm{cons}$ is used to update the CE layer parameters as $\theta_{F_k} \leftarrow \theta_{F_k} - \eta \nabla_{\theta_{F_k}}L_\mathrm{cons}$, where $\eta$ denotes the learning rate.

In this work, we design $L_\mathrm{cons}$ based on the NT-Xent loss used in SimCLR~\cite{chen2020simple}, a representative contrastive learning method. However, unlike conventional contrastive learning, where a pair of embeddings is constructed for each input, our approach generates $K$ embeddings—one from each device—for the same input, yielding $\frac{K(K-1)}{2}$ possible pairs. An appropriate contrastive loss function must therefore be defined over these pairwise relationships. A straightforward approach would be to compute the standard contrastive loss for all possible pairs and take their average. However, this leads to a computational complexity of $\mathcal{O}(K^2)$, which is impractical in large-scale settings. 

To address this issue, we adopt an approximate strategy in which the average embedding across all devices—acting as a centroid in the embedding space—is used as a reference, and each device’s embedding is encouraged to align with this average. Specifically, the loss $L_\mathrm{cons}$ is defined as:
\begin{equation}
L_\mathrm{cons} = \frac{1}{K}\sum_{k=1}^{K} \frac{1}{2N} \sum_{x \in B} \left[ l(z_{k,x}', z_x^\mathrm{avg}) + l(z_x^\mathrm{avg}, z_{k,x}') \right],
\label{eq:loss_consensus}
\end{equation}
where $K$ is the number of devices, $B$ is a mini-batch of size $N$, and $z_x^\mathrm{avg}$ denotes the average embedding across all devices for sample $x$, computed as
$
z_x^\mathrm{avg} = \frac{1}{K} \sum_{k=1}^{K} z_{k,x}'.
$
By replacing explicit pairwise comparisons with alignment to a shared centroid, this approximation reduces the computational complexity to $\mathcal{O}(K)$ while still promoting feature harmonization across devices using only unlabeled data.

For embeddings $z_{i,x}$ and $z_{j,x}$, the function $l(z_{i,x}, z_{j,x})$ is defined as:
\begin{equation}
l(z_{i,x}, z_{j,x}) = -\log\frac{\exp(\mathrm{sim}(z_{i,x}, z_{j,x})/\tau)}{\sum_{x' \in B \setminus \{x\}} \exp(\mathrm{sim}(z_{i,x}, z_{j,x'})/\tau)},
\end{equation}
where $\tau$ is a temperature parameter, and $\mathrm{sim}(\bm{u}, \bm{v})$ denotes cosine similarity, defined as $\mathrm{sim}(\bm{u}, \bm{v}) = \bm{u}^\top \bm{v} / (\lVert \bm{u} \rVert \cdot \lVert \bm{v} \rVert)$. 
In this formulation, the numerator term encourages embeddings $z_{i,x}$ and $z_{j,x}$, produced by devices $i$ and $j$ for the same input $x$, to be pulled closer together. The denominator contrasts $z_{i,x}$ with embeddings $z_{j,x'}$ from different samples $x' \ne x$ in the batch, thus pushing them apart. 

In practical training, one of the participating devices temporarily serves as an aggregator. It collects the consensus embeddings from all devices, computes $L_\mathrm{cons}$, and returns to each device the embedding-level gradient $\frac{\partial L_\mathrm{cons}}{\partial z_{k,x}'}$ associated with that device's consensus embedding.
Finally, each device backpropagates the received embedding-level gradient through its local CE layer $F_k$, computes the gradient $\nabla_{\theta_{F_k}}L_\mathrm{cons}$ locally, and updates its CE layer parameters via
$
\theta_{F_k} \leftarrow \theta_{F_k} - \eta \nabla_{\theta_{F_k}}L_\mathrm{cons},
$
where $\eta$ is the learning rate. During this process, the parameters $\theta_{M^\mathrm{H}_k}$ of the pretrained head network $M^\mathrm{H}_k$ remain fixed, i.e., $\nabla_{\theta_{M^\mathrm{H}_k}}L_\mathrm{cons} = 0$.

\subsection{Unsupervised Training of Cooperative Output Layer}

\begin{figure}
    \centering
    \includegraphics[width=\linewidth]{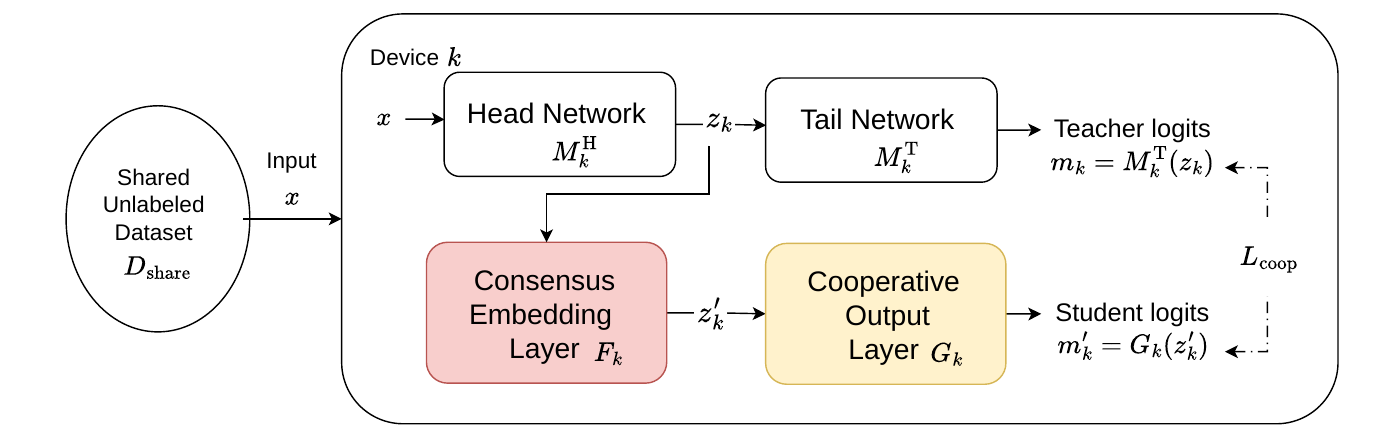}
    \caption{Overview of the CO layer training procedure. Each device generates a teacher logit from its own head and tail networks, and a student logit from the consensus embedding. The CO layer is updated to align the student logit with the teacher logit.
}
    \label{fig:study-method-cooperative}
\end{figure}

Fig.~\ref{fig:study-method-cooperative} illustrates the overall training process for the CO layer. 
The CO layer predicts class scores from the consensus embedding $z'_k$. Inspired by knowledge distillation, we use the output logits $m_k$ from the device’s own pretrained model $M_k$ as teacher logits to guide the training of the CO layer without requiring ground-truth labels.

For each input $x \in D_\mathrm{share}$, device $k$ first generates the teacher logits $m_{k,x} = M_k(x)$ using its own pretrained head and tail networks, and simultaneously obtains the consensus embedding $z_{k,x}' = F_k(z_{k,x})$. The CO layer $G_k$ then produces the student logits $m_{k,x}' = G_k(z_{k,x}')$, which are trained to match the teacher logits using a Kullback–Leibler (KL) divergence-based distillation loss:
\begin{equation}
L_\mathrm{coop} = T^2 \sum_{x \in B} \mathrm{KL}\left( \mathrm{softmax}\left(\frac{m_{k,x}'}{T} \right) \bigg|\bigg| \mathrm{softmax}\left(\frac{m_{k,x}}{T} \right) \right),
\end{equation}
where $T$ is the temperature parameter and $B$ is the mini-batch. The factor $T^2$ compensates for the gradient scaling introduced by temperature-softened logits, following~\cite{hinton2015distilling}. The CO layer’s parameters $\theta_{G_k}$ are updated via gradient descent as $\theta_{G_k} \leftarrow \theta_{G_k} - \eta \nabla_{\theta_{G_k}} L_\mathrm{coop}$, while the parameters of the pre-trained model $M_k$ and CE layer $F_k$ are kept fixed, i.e., $\nabla_{\theta_{M_k}} L_\mathrm{coop} = 0$ and $\nabla_{\theta_{F_k}} L_\mathrm{coop} = 0$. This ensures that only the CO layer learns to translate consensus embeddings into accurate class predictions.

Through this approach, the CO layer learns to predict distributions consistent with those of the pretrained model for each sample, thereby enabling predictions using only unlabeled data.

\subsection{Details of Inference Process and Energy-based Ensemble}
\label{subsec:inference}
The inference process in CE-FI proceeds as follows. When any device $k$ receives an input $x^{(k)}$, it generates the intermediate feature $z_k = M^\mathrm{H}_k(x^{(k)})$ using the head network of its locally deployed pretrained model. This intermediate feature $z_k$ is then mapped into the common embedding space through the device's CE layer, yielding the consensus embedding $z'_k = F_k(z_k)$, which is then shared with other devices. Each recipient device $k'$ predicts the output logits $m'_{k'} = G_{k'}(z'_k)$ using its own CO layer.
Finally, the predictions obtained from multiple devices, represented as output logits ${m'_1, m'_2, \dotsc, m'_K}$, are gathered by device $k$ and combined to produce the final prediction.

However, since each device is trained on a different class distribution, the input $x$ may belong to an unseen class for some models, leading to unreliable logits $m'_{k'}$. At inference time, the system does not know in advance which devices are reliable for the current input, so it cannot directly determine which predictions should be trusted. 
Therefore, designing an effective ensemble strategy to aggregate ${m'_{k'}}$ is essential for achieving robust inference.

Ensemble strategies are broadly classified into aggregation-based and selection-based methods.  
Aggregation-based methods combine outputs from multiple devices to improve robustness. Among them, hard voting and soft voting are widely used \cite{gandhi2015hard,sherazi2021soft}.  
In hard voting, each device predicts a label $y_k^*$, and the final output is determined by majority vote:  
$y^* = \mathrm{mode}(\{y_k^*\}_{k=1}^K)$.  
Soft voting averages softmax probabilities:  
$\overline{p} = \frac{1}{K} \sum_{k=1}^K \mathrm{softmax}(m_k')$, $y^* = \arg \max_i \overline{p}_i$.  
Similar to soft voting, Logits Averaging computes the mean of logits:  
$\overline{m} = \frac{1}{K} \sum_{k=1}^K m_k'$, $y^* = \arg \max_i \mathrm{softmax}(\overline{m})_i$.
These aggregation-based methods can work well when most devices provide reliable predictions. For example, when the class distributions across devices are relatively consistent and the degree of label mismatch is small, aggregating outputs from multiple devices can improve robustness by averaging out individual errors.  However, under severe label mismatch, the input may belong to an unseen class for many devices. In such cases, aggregation may combine unreliable logits from multiple devices, introducing noise and potentially degrading the final prediction.

Selection-based methods instead choose the most reliable device output by evaluating the confidence of each device’s prediction.  
A simple approach is the Maximum Softmax method, which selects the device with the highest softmax value:  
$k^* = \arg \max_k \max_i \mathrm{softmax}(m_k')_i$, $y^* = \arg \max_i \mathrm{softmax}(m_{k^*}')_i$.  
The Minimum Entropy method chooses the device with the lowest entropy, where $p_k = \mathrm{softmax}(m_k')$ and $H(p_k) = -\sum_i p_{k,i} \log p_{k,i}$:  
$k^* = \arg \min_k H(p_k)$, $y^* = \arg \max_i p_{k^*,i}$. 
However, since these methods rely on softmax probabilities, they may still exhibit overconfident behavior for out-of-distribution (OOD) samples.

To mitigate this limitation, recent studies have explored energy-based scores as an alternative confidence measure for OOD detection~\cite{liu2020energy}. 
The energy score is defined as  
\begin{equation}
E(m_k') = -\log\left(\sum_i \exp{m'_{k,i}}\right),
\end{equation}  
which quantifies uncertainty directly from the logits without relying on softmax. Unlike softmax-based scores, the energy score is less affected by probability normalization, which can produce artificially confident predictions even for unseen classes.
Based on this score, the Minimum Energy method selects the device with the lowest energy value and outputs the label corresponding to the highest logit of that device:
$k^* = \arg \min_k E(m_k')$, $y^* = \arg \max_i \mathrm{softmax}(m_{k^*}')_i$.
Since the energy score has been shown to correlate well with model uncertainty in OOD detection settings, it is likely to suppress unreliable predictions produced by devices that have not observed the corresponding classes, particularly under severe label mismatch.  
We therefore use Minimum Energy as the default ensemble rule in CE-FI 
Sect.~\ref{sec:evaluation} provides an ablation study comparing different ensemble strategies and analyzes their performance under various non-IID conditions.

\subsection{Theoretical Relationship between CE-FI and Input-sharing FI}
\label{sec:theoretical-analysis}

The proposed CE-FI framework enables cooperative inference among multiple pretrained models under strict constraints, where neither input data, model parameters, nor a common encoder can be shared across devices. This setting represents one of the most restrictive forms of Federated Inference (FI), as no common input or representation space is assumed. To better understand the performance implications of these constraints, we consider a relaxed reference setting in which all devices are allowed to observe the same input and directly aggregate their prediction outputs. 
We refer to this configuration as Input-sharing FI.

This section formally clarifies the relationship between CE-FI and Input-sharing FI, and aims to provide theoretical insight into the factors that limit the performance of CE-FI.

\subsubsection{Equivalence under Ideal Consensus Embedding and Cooperative Output Layers}

We first consider an idealized setting in which both the CE layer and the CO layer operate perfectly. Specifically, we assume that for any input sample $x$, the CE layer $F_k$ of every device $k$ produces an identical consensus embedding, i.e., $z_{i,x}'=z_{j,x}', \forall i,j$.  

Under this assumption, for any device $i$, the output of its CO layer for a consensus embedding shared by another device $j$ coincides with that for its own embedding, namely,
$G_i(z_{j,x}') = G_i(z_{i,x}')$. 
Furthermore, suppose that the CO layer is ideally trained using a knowledge-distillation-based loss and can exactly reproduce the output logits of the original local model. Then, the cooperative distillation loss satisfies
\begin{equation}
\begin{aligned}
    L_\mathrm{coop} &= T^2 \sum_{x \in B} \mathrm{KL}\left( \mathrm{softmax}\left(\frac{m_{k,x}'}{T} \right) \bigg|\bigg| \mathrm{softmax}\left(\frac{m_{k,x}}{T} \right) \right) \\ &=0,
\end{aligned}
\end{equation}
which implies
\begin{equation}
    \mathrm{softmax}\left(\frac{m_{k,x}'}{T} \right) = \mathrm{softmax}\left(\frac{m_{k,x}}{T} \right).
\end{equation}
Since the softmax function is shift-invariant, it follows that $\frac{m_{k,x}'}{T} = \frac{m_{k,x}}{T} + c_{k,x}$, 
where $c_{k,x}$ is a constant vector whose elements are identical, depending on sample $x$ and device $k$. Consequently,
$m_{k,x}'=m_{k,x}+Tc_{k,x}$. 
Letting $p_{k,x}=\mathrm{softmax}(m_{k,x})$ and $p_{k,x}'=\mathrm{softmax}(m_{k,x}')$, we obtain $
    p_{k,x}=p_{k,x}'$.

That is, the prediction probability distribution produced by CE-FI at each device exactly matches that of the corresponding pretrained local model. Therefore, when the same ensemble method $\mathcal{A}$ is applied in both Input-sharing FI and CE-FI, for any input $x$, let $p_{k,x}'^n = \mathrm{softmax}(m_{k,x}'^n)$. Then, we have 
\begin{equation}
    \mathcal{A}(\{p_{k,x}\}_{k=1}^K)=\mathcal{A}(\{p_{k,x}'^n\}_{k=1}^K),
\end{equation}
which implies that the predicted labels and inference accuracy are identical.
Note that when ensemble methods operating directly on logits are used, equivalence between Input-sharing FI and CE-FI requires the ensemble function to be invariant to constant shifts.

\subsubsection{Effect of Consensus Embedding Errors}

We next consider a more realistic setting in which consensus embeddings are not perfectly aligned, and discrepancies remain across devices. Specifically, for any pair of devices $i,j$, we assume that there exists a sufficiently small constant $\epsilon$ such that $\lVert z_{i,x}' - z_{j,x}'\rVert \leqq \epsilon$. 

Suppose that the CO layer is implemented as a two-layer fully connected network, which can be expressed as $G_i(x)=W_2\mathrm{ReLU}(W_1x+b_1)+b_2$. Since $ \lVert \mathrm{ReLU}(u)-\mathrm{ReLU}(v) \rVert \leqq \lVert u- v \rVert$, we obtain the following bound:
\begin{equation}
    \begin{aligned}
\lVert G_i(z_{i,x}')-G_i(z_{j,x}') \rVert \leqq \lVert W_2\rVert \lVert W_1 \rVert \epsilon.
\end{aligned}
\end{equation}
Let $L=\lVert W_2\rVert \lVert W_1\rVert$.
Then, by exploiting the Lipschitz continuity of the softmax function, we can be bound the discrepancy between $p_i'=\mathrm{softmax}(m_{i}'))$ and $p_i'^j = \mathrm{softmax}(m_i'^j)$ as
\begin{equation}
    \begin{aligned}
\lVert p_i' - {p_i^j}' \rVert &\leqq \lambda \lVert (m_i'-m_i'^j) \rVert \\ &\leqq L \lambda\epsilon.
\end{aligned}
\end{equation}

From the previous discussion, when the CO layer operates ideally, we have $p_{k,x}=p_{k,x}'$. Therefore, if the same ensemble method $\mathcal{A}$ is applied in both Input-sharing FI and CE-FI, then for any input $x$,
\begin{equation}
    \mathcal{A}(\{p_k\}_{k=1}^K)=\mathcal{A}(\{p_k'\}_{k=1}^K).
\end{equation}
In order for CE-FI and Input-sharing FI to achieve identical accuracy, it suffices for $\epsilon$ to be small enough that the predicted labels remain unchanged, i.e., $\mathrm{argmax}(p_{k}') = \mathrm{argmax}(p_{k}'^n)$. More concretely, let $(p_i'^j)^\mathrm{1st}$ and $(p_i'^j)^\mathrm{2nd}$ denote the largest and second-largest elements of $p_i'^j$, respectively. If
\begin{equation}
    |(p_i'^j)^\mathrm{1st}-(p_i'^j)^\mathrm{2nd}| > 2L \lambda \epsilon,
\end{equation}
then the prediction label does not change between $p_i'$ and $p_i'^j$, and the inference accuracy coincides.

In the experimental settings considered in this work, these sufficient conditions are not strictly satisfied. Nevertheless, this analysis theoretically indicates that CE-FI can approach the performance of Input-sharing FI as the quality of the CE layer and the CO layer improves.
In the next section, we empirically compare the inference accuracy of CE-FI with Input-sharing FI. Moreover, through an ablation study in Sect.~\ref{sec:bottleneck}, we investigate how improving embedding alignment—e.g., via consensus space unification—affects practical performance and validates the implications of this theoretical characterization.
\section{Performance Evaluation}
\label{sec:evaluation}
\subsection{Setup}
\label{sec:setup}
 We assume a cross-silo scenario involving a small number of devices—3 in the primary evaluations and up to 10 in selected experiments—each possessing a model trained on its own local dataset while sharing a common task. The CE and CO layers are trained using a shared unlabeled dataset, enabling cooperation among the pretrained models to improve inference accuracy. As this paper focuses on the proof of concept for CE-FI, we assume ideal conditions where differences in computational capacity, communication bandwidth, and resulting delays or transmission losses between devices are negligible.

\vspace{2mm}
\noindent\textbf{Task and Dataset:}
To validate the effectiveness of CE-FI across multiple modalities, we consider three types of tasks: image classification, text classification, and time-series classification.

\noindent\textbf{Image classification.}
We evaluate CE-FI on two benchmark datasets, CIFAR-10 and CIFAR-100. CIFAR-10 is a standard image classification dataset consisting of 10 classes, with 50,000 training samples and 10,000 test samples. CIFAR-100 consists of 100 fine classes, which are further grouped into 20 coarse classes, and contains the same number of samples as CIFAR-10. In this work, we focus on the more challenging setting of CIFAR-100 coarse-class classification. For each dataset, we allocate 80\% of the training samples to device-local datasets according to the method described later and use them to pretrain device-specific local models. We then discard the labels of the remaining 20\% and use the resulting data as a shared unlabeled dataset to train the CE and CO layers.

\begin{figure}
    \centering
    
    \subfloat[Mild\label{fig:label-distribution-mild}]{
    \centering
    \includegraphics[width=0.45\linewidth]{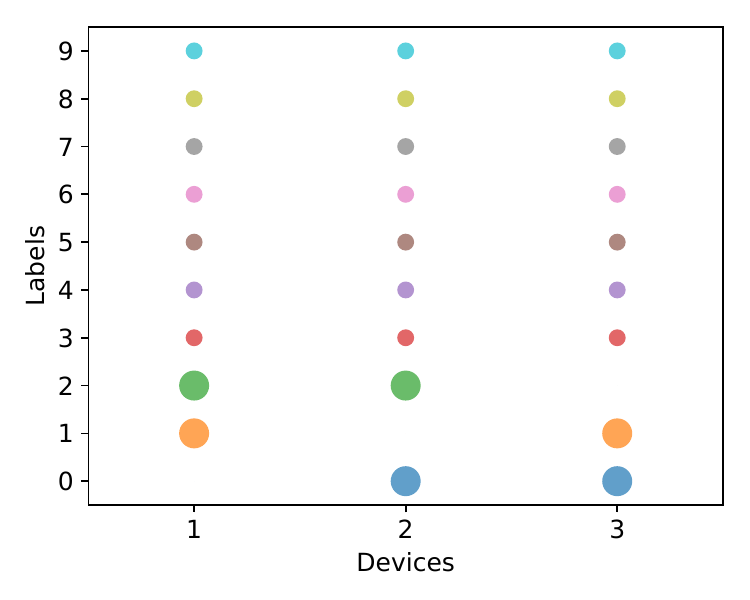}}
    \hfill
    \subfloat[Moderate\label{fig:label-distribution-moderate}]{
    \centering
    \includegraphics[width=0.45\linewidth]{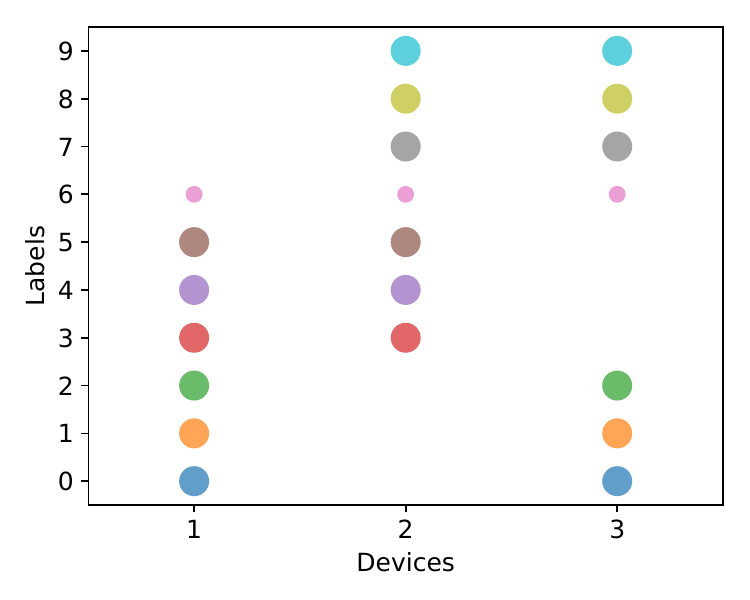}}
    \hfill
    \subfloat[Skewed\label{fig:label-distribution-skewed}]{
    \centering
    \includegraphics[width=0.45\linewidth]{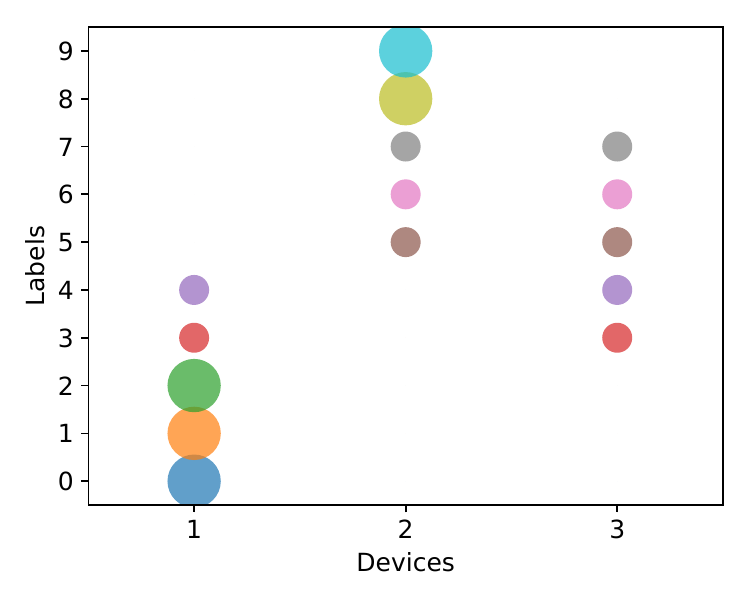}}
    \hfill
    \subfloat[Disjoint\label{fig:label-distribution-disjoint}]{
    \centering
    \includegraphics[width=0.45\linewidth]{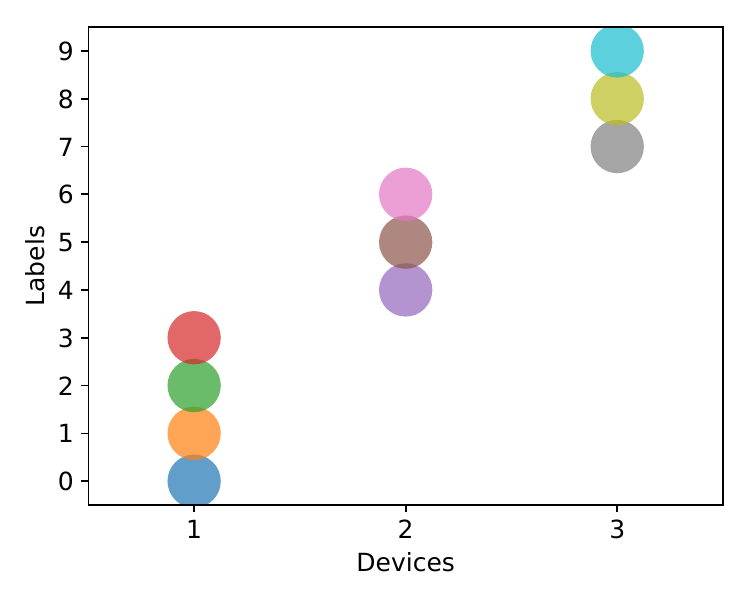}}
\caption{Distribution of labels in the 3-device setting: (a)Mild (b)Moderate (c)Skewed (d)Disjoint. Dot size indicates the number of samples for each label.}
\label{fig:label-distribution}
\end{figure}

To evaluate CE-FI under diverse non-IID conditions, we constructed local datasets with imbalanced label distributions using two partitioning methods: (1) a manual split with controlled label overlap among devices, and (2) a Dirichlet-based random split, commonly used in FL evaluations. 
In the manual split setting, local datasets were partitioned into four levels of non-IIDness:  
(a) Mild — each device is missing one label;  
(b) Moderate — each label is shared by roughly half the devices without extreme imbalance;
(c) Skewed — certain labels exist on only a few devices;  
(d) Disjoint — each device holds a completely disjoint label set.  
An example for CIFAR-10 with three devices is shown in Fig.~\ref{fig:label-distribution}. For CIFAR-100, we apply the same partitioning scheme to fine classes within each coarse class, introducing heterogeneity in the underlying feature distributions across devices.

In the Dirichlet-based setting, each device's label distribution is sampled from a Dirichlet distribution with $\alpha \in \{0.1, 0.5\}$, where smaller $\alpha$ results in more skewed splits.  
For CIFAR-100, we consider two variants depending on the partitioning granularity: (a) CIFAR-100-fine, where partitioning is performed over fine classes, and
(b) CIFAR-100-coarse, where partitioning is performed over coarse classes. In both cases, the classification task remains coarse-class prediction. We repeat each experiment five times to ensure robustness.

As an ablation study, we also evaluate a scratch setting without pretrained encoders. In this setting, we use MNIST, Fashion MNIST, and CIFAR-10, and conduct experiments under both manual and Dirichlet partitioning, as well as scenarios with an increased number of devices. Details are provided in Sect.~\ref{sec:scratch}.

\noindent\textbf{Text classification.}
For the text task, we conduct experiments on the IMDB dataset. While IMDB is commonly used for binary sentiment classification (positive/negative)~\cite{maas-etal-2011-imdb}, we adopt a more challenging multi-class setting in which the task is to classify movie genres. Following a publicly available IMDB genre-annotated benchmark~\cite{kaggle_imdb_genre}, we evaluate a 27-class genre classification task. The dataset contains approximately 54,000 training samples and exhibits substantial class imbalance across genres. As in the image classification setting, the training data are partitioned into device-local datasets and a shared unlabeled subset.

To capture diverse non-IID conditions, we design three label-partition settings while keeping the total number of samples per device approximately equal. Since class frequencies are imbalanced, we categorize classes by sample size: large ($\geq$5000), medium (1000--5000), and small ($<$1000), and design splits accordingly: (a) Anchored: all devices share a fraction of samples per class, while the remaining samples are skewed toward designated devices. In particular, medium and large classes are assigned to two devices, and small classes to one device. In our experiments, each device receives 15\% of samples from every class as anchor data, while the remaining 55\% are allocated to designated devices; we denote this setting as Anchored-15.
(b) Balanced: samples are distributed across multiple devices depending on class size (large to three devices, medium to two, small to one).
(c) Disjoint: each class is held by only a single device.

\noindent\textbf{Time-series classification.}
For the time-series task, we evaluate on the UCI Human Activity Recognition (UCI-HAR) dataset, a representative benchmark for human activity recognition using smartphone accelerometer and gyroscope signals. We use only the accelerometer signals to demonstrate the effectiveness of CE-FI on time-series modalities under a simplified sensor setting. The dataset contains approximately 7,000 training samples and 3,000 test samples. The task is to classify six activities: walking, walking upstairs, walking downstairs, sitting, standing, and lying.  We adopt the same manual partitioning scheme as in image classification.

\vspace{2mm}
\noindent\textbf{Model Architecture:}
In all settings, each device maintains a local model $M_k$, which consists of a head network that maps an input sample into an intermediate feature representation, and a tail network that produces task-specific predictions based on this representation. While the head network varies depending on the task and experimental configuration, the tail network is common to all settings. Specifically, we employ a simple two-layer fully connected classifier with 1024 units that takes the head output as input, using ReLU activations and a dropout rate of 0.3.

For image classification with pretrained encoders, we consider two configurations to evaluate the applicability of the proposed framework: (a) Unified-architecture (ViT)  setting: all devices adopt the same Vision Transformer (ViT) architecture. To assess representation differences induced by pretraining objectives, each device adopts a distinct pretrained model: supervised pretraining on ImageNet~\cite{imagenet-2015}, image–text contrastive pretraining via CLIP~\cite{clip-pmlr-v139-radford21a}, and self-supervised pretraining via DINOv3~\cite{siméoni2025dinov3}.
(b) Heterogeneous-architecture setting: devices employ different encoder architectures, namely an ImageNet-supervised ViT, ResNet-50~\cite{resnet-7780459}, and ConvNeXtV2~\cite{convnextv2-10205236}. 
In the scratch setting, we design dataset-specific architectures. For MNIST and Fashion MNIST, the head network is implemented as a three-layer convolutional neural network (CNN), where each layer consists of a $5\times5$ convolution, ReLU activation, and max pooling, followed by batch normalization after the third layer. For CIFAR-10, we adopt a standard ResNet-18 architecture.
For the text and time-series tasks, we train task-specific head networks by considering the input structure and data characteristics: an LSTM-based encoder for text classification and a hybrid architecture combining 1D-CNN and GRU for time-series signals.

We designed the CE layer based on SimpleCLIP~\cite{simpleclip2021}, inspired by the vision-language model CLIP~\cite{clip-pmlr-v139-radford21a}, consisting of two fully connected layers with ReLU activation and dropout, including skip connection and layer normalization. The dimension of the consensus embedding is set to 256 in all experiments as a trade-off between 
representation capacity and communication efficiency. 
The CO layer receives the consensus embedding as input and produces final logits through two fully connected layers with ReLU activation and a dropout rate of 0.3. 
The primary aim of this study is not to optimize architectural design but to test whether cooperative inference is possible without the sharing assumptions used in conventional FI. Therefore, we adopt a simple structure, acknowledging potential future improvements.

\vspace{2mm}
\noindent\textbf{Training Procedure:}
All models are trained using the Adam optimizer with a learning rate of $10^{-3}$ and weight decay of $5\times10^{-4}$. 
A cosine annealing learning rate scheduler is applied, where the learning rate gradually decreases to $10^{-5}$ over the training process.
As CE-FI involves multiple training stages, we use different batch sizes and training epochs depending on the stage and experimental setting.

\noindent\textbf{Pretrained encoder setting.}
Each device first trains its task-specific tail network while keeping the pretrained encoder fixed. 
The tail networks are trained with a batch size of 64 for up to 20 epochs with early stopping. 
For early stopping, 20\% of the local training data is used as validation data, and training stops when the validation loss does not improve for five consecutive epochs.
Next, the CE layer is trained using the shared unlabeled dataset with a batch size of 512 for 100 epochs. 
Finally, the CO layer is trained for 20 epochs with a batch size of 64.

\noindent\textbf{Scratch setting.}
When training from scratch, the head and tail networks are jointly trained using a batch size of 128 for up to 100 epochs with the same early stopping criterion. The CE layer is then trained under the same configuration as in the pretrained encoder setting, and the CO layer is trained for 20 epochs with a batch size of 128.

\vspace{2mm}
\noindent\textbf{Loss Hyperparameters:}
For the CE layer, the temperature parameter in the NT-Xent loss is set to $\tau=0.2$, following the empirical range examined in SimCLR~\cite{chen2020simple} as a moderate value that provides stable contrastive training.
For the CO layer, the distillation temperature is set to $T=3.0$, following empirical observations in the original knowledge distillation study~\cite{hinton2015distilling} that moderate temperatures are effective in practice.

Since the goal of this work is to validate the feasibility of the CE-FI framework rather than to optimize performance through extensive hyperparameter tuning, we select these hyperparameters from representative values and keep them fixed across all experiments.

\vspace{2mm}
\noindent\textbf{Comparison Methods:}
We compare CE-FI with the following baselines for evaluating the performance in multiple experimental scenarios: 
(a) Solo Inference, where each device performs inference using only its own pre-trained model;  
(b) Proposed w/ Oracle Ensemble, an idealized version of CE-FI in which the ensemble always selects the prediction from the device that provides the correct label, if available.
(c) Input-sharing FI, which relaxes the constraints on input sharing—each device receives the same input, and the final prediction is obtained via soft voting across their outputs.
(d) Edge Ensemble, an alternative FI configuration where devices operate under a shared pretrained encoder that produces aligned intermediate representations, and the final prediction is similarly aggregated via soft voting.

In addition, an ablation study compares various ensemble strategies described in Sect.~\ref{subsec:inference}, including soft voting, Logits Averaging, Maximum Softmax, Minimum Entropy, Minimum Energy, and a naive Random Selection method.

\subsection{Baseline Comparison}
\begin{figure}[htbp]
  \centering

  \subfloat[CIFAR-10, Unified-architecture (ViT)\label{fig:baseline-cifar10-vit}]{
    \centering
\includegraphics[width=\linewidth]{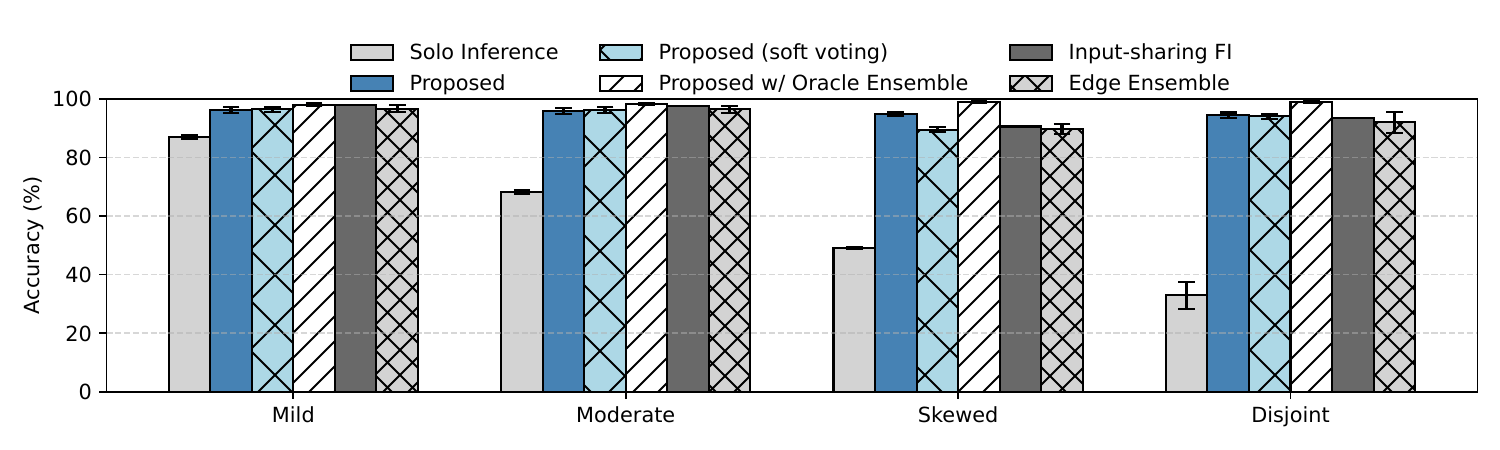}
  }
  \hfill
  \subfloat[CIFAR-10, Heterogeneous-architecture\label{fig:baseline-cifar10-hetero}]{
    \centering
\includegraphics[width=\linewidth]{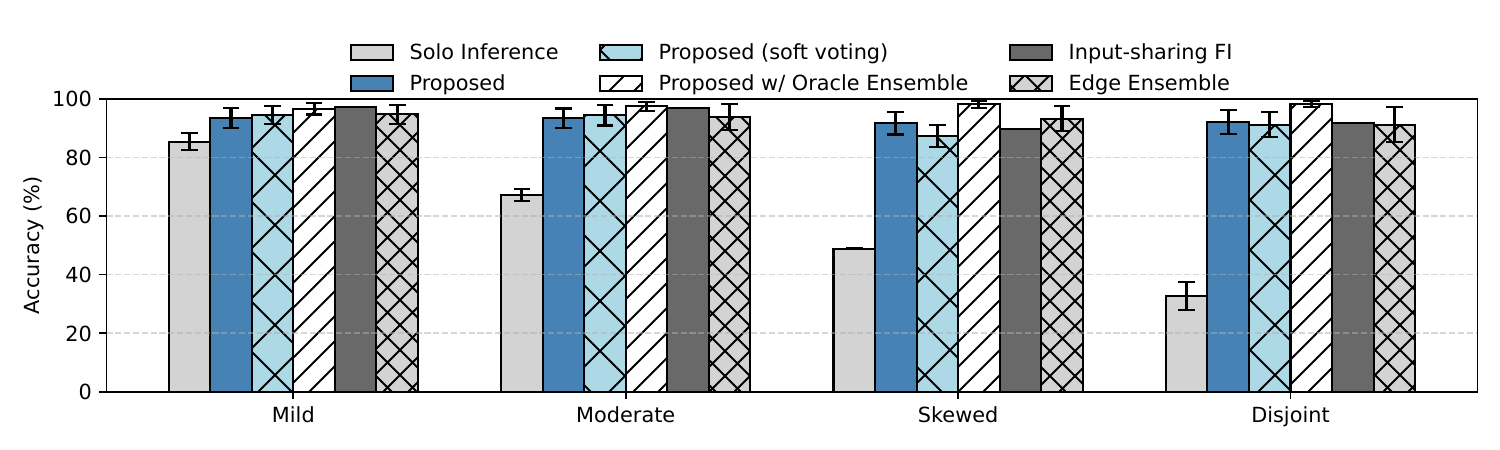}
  }
  \hfill
  \subfloat[CIFAR-100, Unified-architecture (ViT)\label{fig:baseline-cifar100-vit}]{
    \centering
    \includegraphics[width=\linewidth]{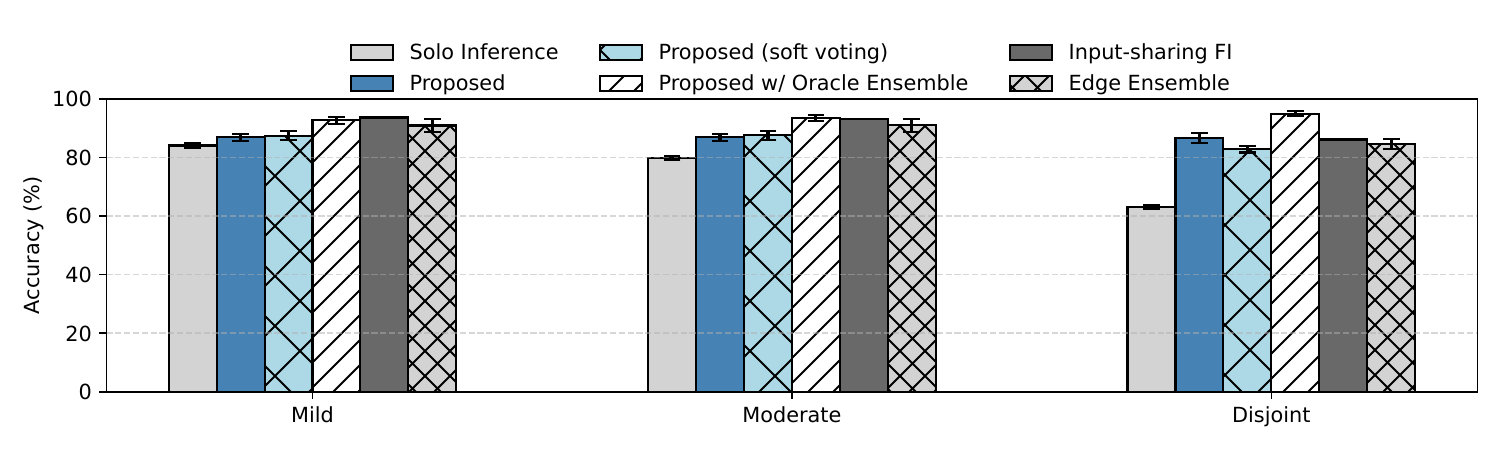}
  }
  \hfill
  \subfloat[CIFAR-100, Heterogeneous-architecture\label{fig:baseline-cifar100-hetero}]{
    \centering
    \includegraphics[width=\linewidth]{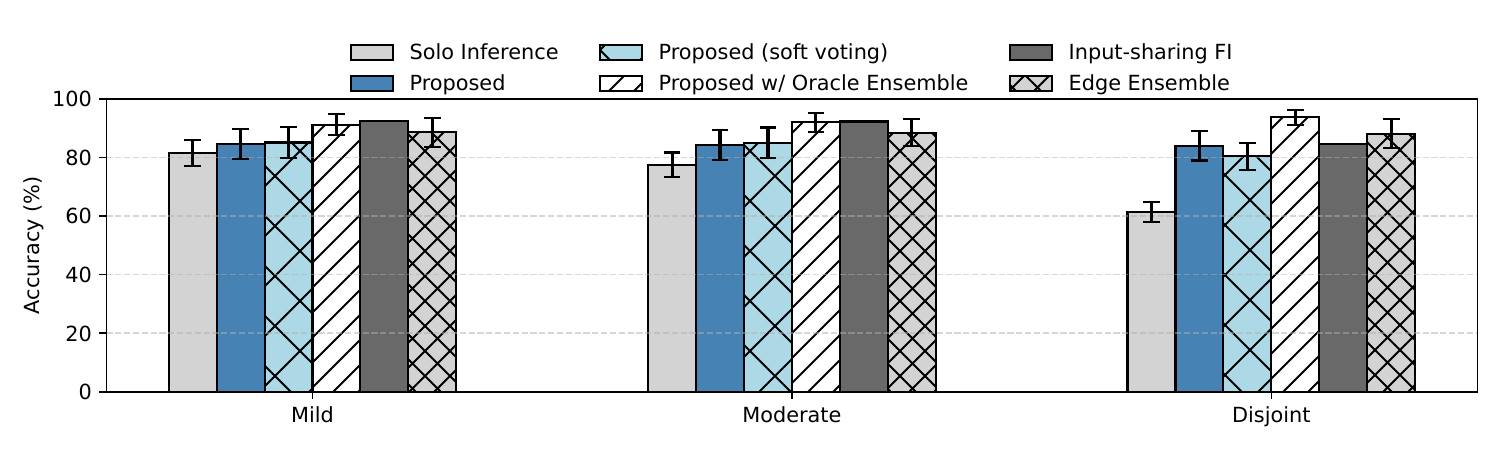}
  }
  
  \caption{Accuracy across label partition strategies under different datasets and architectural settings (3 devices). (a) CIFAR-10 / Unified-architecture (ViT), (b) CIFAR-10 / Heterogeneous-architecture, (c) CIFAR-100 / Unified-architecture (ViT), (d) CIFAR-100 / Heterogeneous-architecture.}
  \label{fig:baseline-comparison}
\end{figure}

Fig.~\ref{fig:baseline-comparison} shows the inference accuracy on CIFAR-10 and CIFAR-100 classification tasks under settings with pretrained encoders. For each dataset, evaluations were conducted under multiple non-IID label partition conditions described in Sect.~\ref{sec:setup}. For a fair comparison with FI-style baselines, we report results for both CE-FI with Minimum Energy and CE-FI with soft voting.
Overall, CE-FI consistently outperforms Solo Inference across all configurations. This improvement is observed not only in relatively mild non-IID settings but also in more challenging scenarios such as Skewed and Disjoint, where label distributions across devices are highly imbalanced.
When comparing CE-FI with conventional FI-based baselines, we focus on the Proposed (soft voting) configuration, which adopts the same output aggregation strategy as Input-sharing FI and Edge Ensemble. Under this setting, CE-FI achieves performance that approaches these FI-based methods across many configurations. 
In addition, using the Minimum Energy selection strategy further improves the performance of CE-FI under certain non-IID conditions, highlighting the importance of the ensemble strategy. 

In the Heterogeneous-architecture setting, where devices employ encoders with different structures, performance varies depending on which device generates the consensus embedding that is shared. This likely reflects differences in how well each encoder aligns with the consensus embedding space. Nevertheless, the average inference accuracy remains consistently higher than Solo Inference under all non-IID conditions, indicating that CE-FI retains a certain degree of robustness against encoder heterogeneity. 
Therefore, CE-FI does not merely compensate for missing classes across devices; it also remains effective when devices learn heterogeneous internal representations.

Nevertheless, although the performance of CE-FI is already close to that of the FI-based baselines in many settings, a consistent gap still remains between the practical ensemble strategies and the Proposed w/ Oracle Ensemble setting. This suggests that further improvements may be possible through more accurate ensemble selection mechanisms.

\begin{figure}[t]
  \centering

  \subfloat[CIFAR-100, individual feature (only ImageNet supervised)\label{fig:baseline-cifar100-embedding-imagenet-encoder}]{
  \centering
    \includegraphics[width=0.45\linewidth]{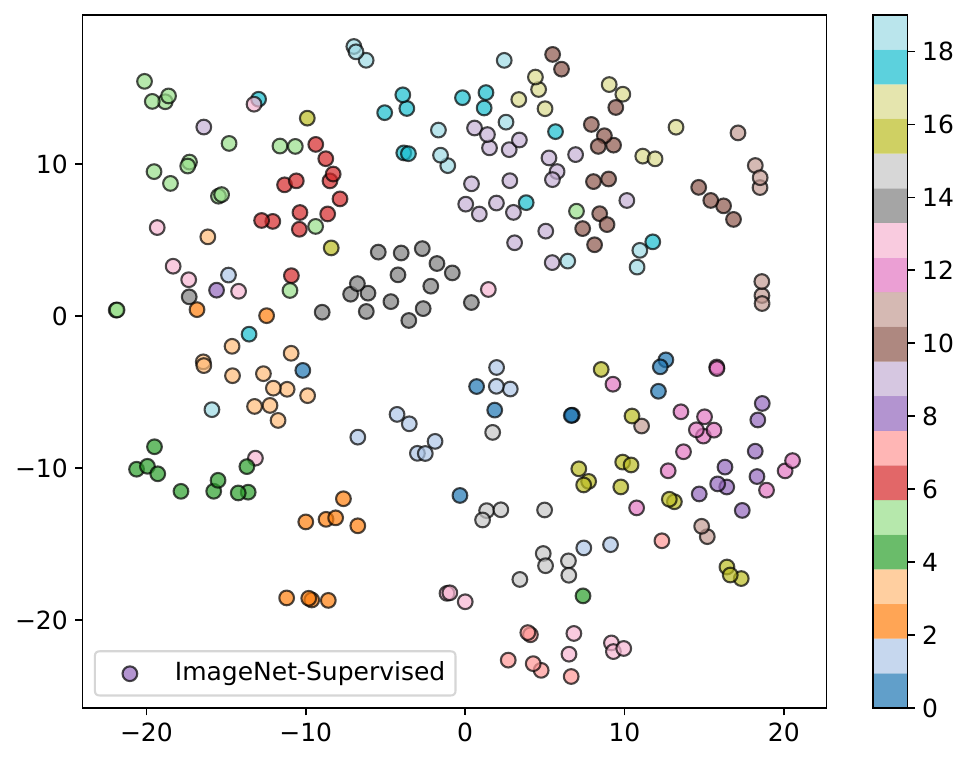}
  }
  \hfill
  
  \subfloat[CIFAR-100, individual feature\label{fig:baseline-cifar100-embedding-ptr-encoder}]{
  \centering
    \includegraphics[width=0.45\linewidth]{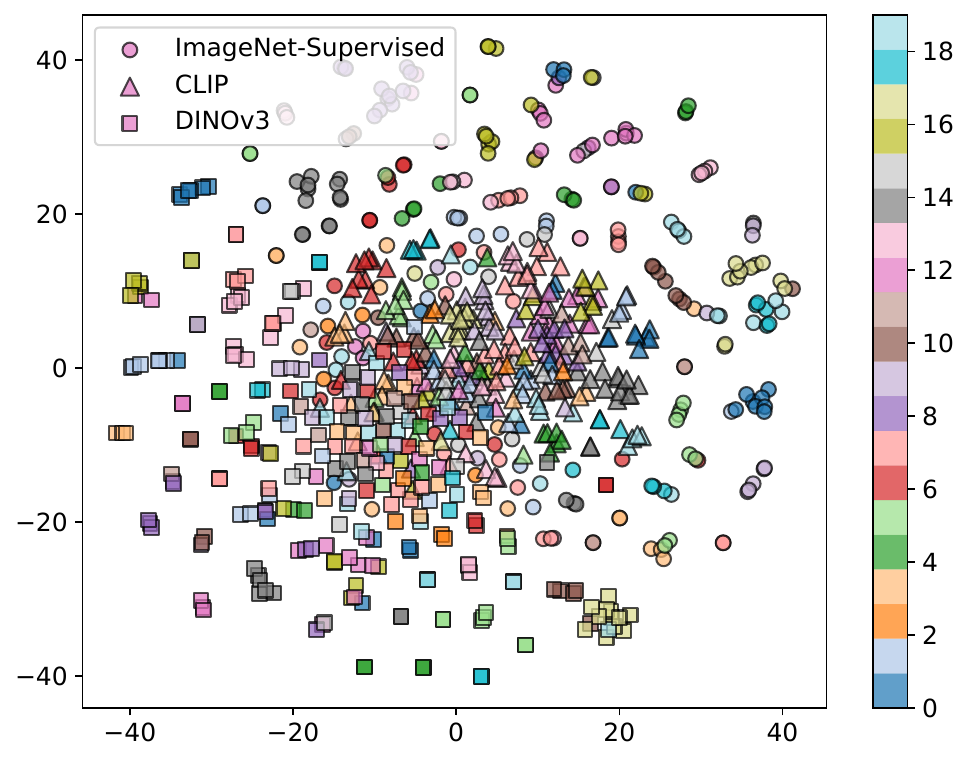}
  }
  \hfill
  \subfloat[CIFAR-100, consensus embedding\label{fig:baseline-cifar100-embedding-consensus}]{
  \centering
    \includegraphics[width=0.45\linewidth]{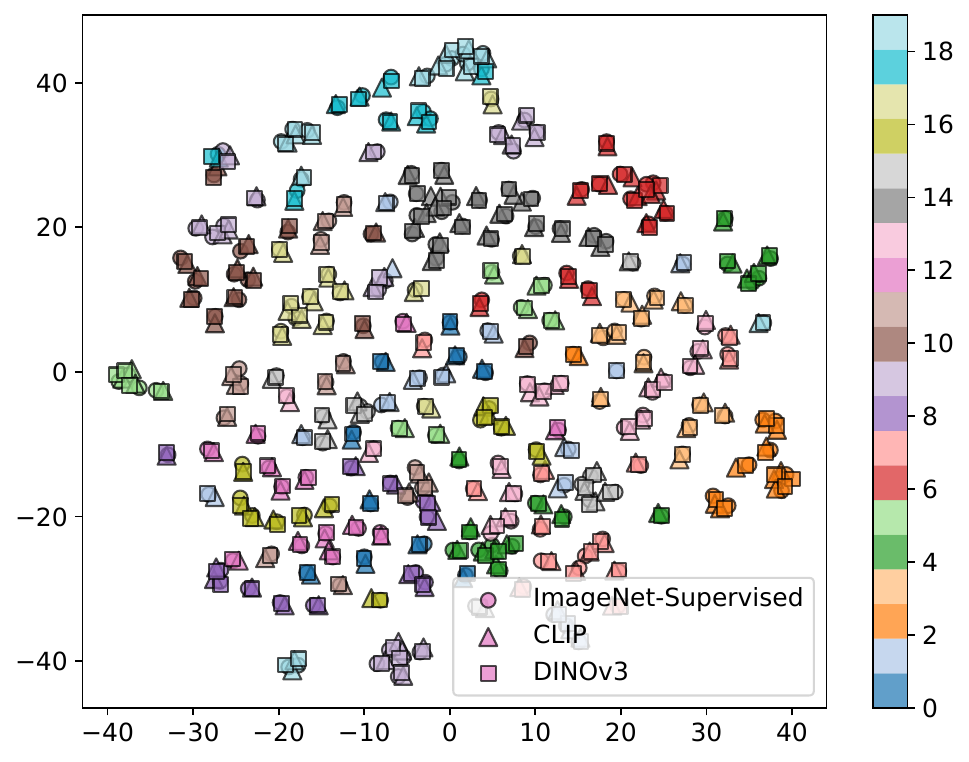}
  }
\caption{t-SNE visualization of feature representations across devices in Unified-architecture setting on CIFAR-100: (a)intermediate features extracted from a single pretrained encoder (ImageNet-supervised). (b)intermediate features from all devices projected into a common space. (c)the consensus embedding space learned by CE-FI, showing enhanced cross-device alignment.}
\label{fig:baseline-embedding}
\end{figure}

To better understand how CE-FI enables cooperative inference
across heterogeneous encoders, we analyze the structure of the
consensus embedding space. Specifically, Fig.~\ref{fig:baseline-embedding} shows t-SNE projections in the Unified-architecture setting on CIFAR-100.
When intermediate features from all devices are projected together, their distributions are highly inconsistent due to differences in encoder architectures and pretraining objectives. As a result, the feature representations from different devices are not directly comparable in a shared space. After passing through the CE layer, however, the resulting consensus embeddings corresponding to the same input sample are mapped to nearby locations across devices.
This indicates that CE-FI achieves a high degree of cross-device consistency in the embedding space. Moreover, a coarse class-wise clustering structure is also observed, suggesting that the consensus embedding space preserves semantic information useful for inference rather than performing mere feature matching.
These observations suggest that CE-FI successfully constructs a shared representation space that preserves semantic structure across devices, enabling consistent prediction behavior across devices. Such representational consistency likely contributes to the strong inference accuracy observed in the pretrained encoder setting.

\subsection{Multimodal Evaluation}
\begin{figure}[htbp]
  \centering

  \subfloat[IMDB\label{fig:baseline-imdb}]{
    \centering
\includegraphics[width=\linewidth]{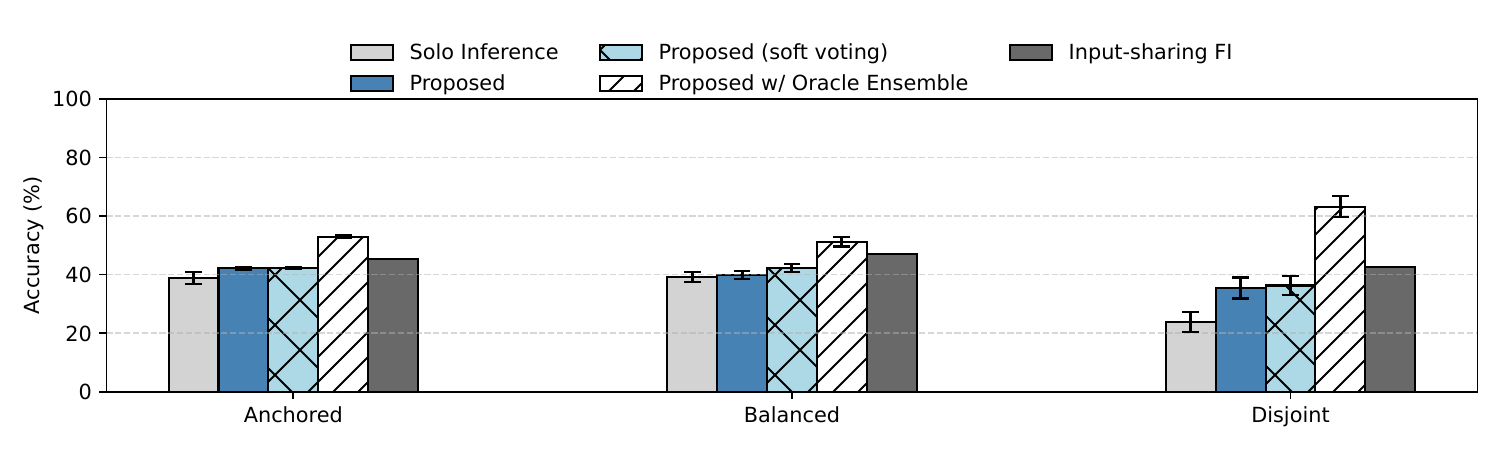}
  }
  \hfill
  \subfloat[UCI-HAR\label{fig:baseline-uci-har}]{
    \centering
\includegraphics[width=\linewidth]{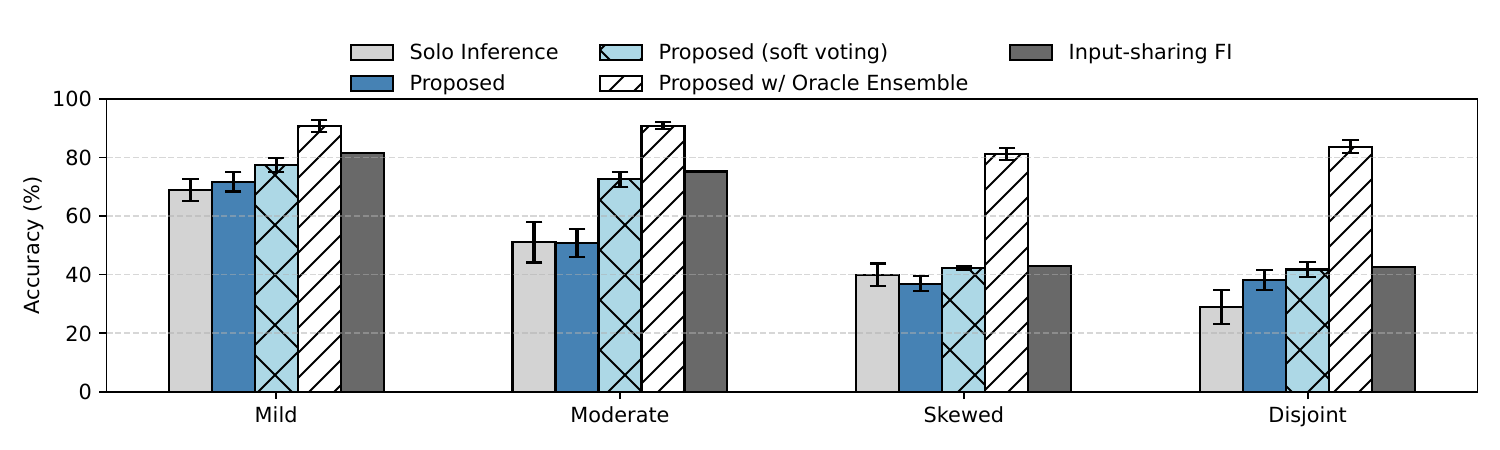}
    }
  \caption{Accuracy across label partition strategies on different modalities (3 devices). (a) IMDB, (b) UCI-HAR.}
  \label{fig:multimodal}
\end{figure}

Fig.~\ref{fig:multimodal} reports the inference accuracy on the IMDB and UCI-HAR datasets under various non-IID label partition conditions. 
These experiments evaluate whether CE-FI can be applied to modalities beyond image classification.
Across both datasets, the performance of CE-FI varies depending on the output aggregation strategy.  In particular, CE-FI with Minimum Energy often provides only marginal improvements over Solo Inference, and in some settings, it does not outperform Solo Inference. However, alternative aggregation strategies, such as soft voting, can yield improved performance in several configurations. This suggests that cooperative inference can provide benefits when an appropriate ensemble strategy is employed. The performance variations are more pronounced in the UCI-HAR dataset, suggesting that the sensitivity to ensemble strategies may depend on  the characteristics of the underlying modality.

Another notable observation is the substantial gap between practical ensemble strategies and the Proposed w/ Oracle Ensemble setting.  This gap likely reflects imperfect ensemble selection and discrepancies in the learned representations across devices. Similar to the image classification results, this indicates that there remains room for improvement.
Overall, CE-FI extends beyond image classification to text and time-series tasks, while the results also highlight the need for modality-aware ensemble design.

\begin{table*}[t]
\centering
\caption{Accuracy of CE-FI with different output ensemble methods (mean $\pm$ std in 3 devices).}
\subfloat[Image Classification - Unified-architecture (ViT)\label{tab:output-ensemble-vit}]{
\setlength{\tabcolsep}{2pt}
\resizebox{\textwidth}{!}{%
\begin{tabularx}{\textwidth}{c|YYYY|YYY}
    \toprule
    \multirow{2}{*}{Method} & \multicolumn{4}{c|}{CIFAR-10} & \multicolumn{3}{c|}{CIFAR-100}  \\
    \cline{2-8}
     & Mild & Moderate & Skewed & Disjoint & Mild & Moderate &  Disjoint  \\
    \midrule
    \shortstack{Random Selection} & \shortstack{$0.8633\pm0.0074$} & \shortstack{$0.6782\pm0.0056$} & \shortstack{$0.4843\pm0.0017$} & \shortstack{$0.3313\pm0.0029$} & \shortstack{$0.8194\pm0.0113$} & \shortstack{$0.7883\pm0.0058$} & \shortstack{$0.6455\pm0.0030$} \\ [1.2ex]
\shortstack{Soft Voting} & \shortstack{$\bm{0.9638}\pm0.0093$} & \shortstack{$\bm{0.9617}\pm0.0093$} & \shortstack{$0.8945\pm0.0077$} & \shortstack{$0.9395\pm0.0081$} & \shortstack{$\bm{0.8733}\pm0.0153$} & \shortstack{$\bm{0.8756}\pm0.0153$} & \shortstack{$0.8280\pm0.0114$}\\  [1.2ex]  
\shortstack{Logits Averaging} & \shortstack{$0.9538\pm0.0096$} & \shortstack{$0.9566\pm0.0092$} & \shortstack{$0.9293\pm0.0077$} & \shortstack{$0.9427\pm0.0092$} & \shortstack{$0.8600\pm0.0154$} & \shortstack{$0.8618\pm0.0153$} & \shortstack{$0.8269\pm0.0132$}\\  [1.2ex]  
\shortstack{Max Softmax} & \shortstack{$0.9594\pm0.0101$} & \shortstack{$0.9506\pm0.0103$} & \shortstack{$0.9406\pm0.0072$} & \shortstack{$0.9399\pm0.0085$} & \shortstack{$0.8650\pm0.0147$} & \shortstack{$0.8665\pm0.0141$} & \shortstack{$0.8632\pm0.0153$} \\  [1.2ex]  
\shortstack{Min Entropy} & \shortstack{$0.9593\pm0.0100$} & \shortstack{$0.9522\pm0.0104$} & \shortstack{$0.9424\pm0.0074$} & \shortstack{$0.9424\pm0.0085$} & \shortstack{$0.8655\pm0.0147$} & \shortstack{$0.8666\pm0.0135$} & \shortstack{$0.8658\pm0.0152$}  \\  [1.2ex] 
\shortstack{Min Energy} & \shortstack{$0.9606\pm0.0097$} & \shortstack{$0.9575\pm0.0093$} & \shortstack{$\bm{0.9477}\pm0.0082$} & \shortstack{$\bm{0.9443}\pm0.0086$} & \shortstack{$0.8683\pm0.0126$} & \shortstack{$0.8688\pm0.0130$} & 
\shortstack{$\bm{0.8668}\pm0.0169$}\\
    \bottomrule
\end{tabularx}}}

\subfloat[Image Classification - Heterogeneous-architecture\label{tab:output-ensemble-hetero}]{
\setlength{\tabcolsep}{2pt}
\resizebox{\textwidth}{!}{%
\begin{tabularx}{\textwidth}{c|YYYY|YYY}
    \toprule
    \multirow{2}{*}{Method} & \multicolumn{4}{c|}{CIFAR-10} & \multicolumn{3}{c|}{CIFAR-100}  \\
    \cline{2-8}
     & Mild & Moderate & Skewed & Disjoint & Mild & Moderate &  Disjoint  \\
    \midrule
    \shortstack{Random Selection} & \shortstack{$0.8513\pm0.0285$} & \shortstack{$0.6681\pm0.0152$} & \shortstack{$0.4808\pm0.0102$} & \shortstack{$0.3266\pm0.0018$} & \shortstack{$0.7924\pm0.0440$} & \shortstack{$0.7578\pm0.0417$} & \shortstack{$0.6142\pm0.0303$} \\ [1.2ex]
\shortstack{Soft Voting} & \shortstack{$\bm{0.9442}\pm0.0314$} & \shortstack{$\bm{0.9429}\pm0.0344$} & \shortstack{$0.8726\pm0.0371$} & \shortstack{$0.9111\pm0.0423$} & \shortstack{$\bm{0.8507}\pm0.0525$} & \shortstack{$\bm{0.8492}\pm0.0526$} & \shortstack{$0.8034\pm0.0463$}\\  [1.2ex]  
\shortstack{Logits Averaging} & \shortstack{$0.9225\pm0.0336$} & \shortstack{$0.9363\pm0.0370$} & \shortstack{$0.8980\pm0.0387$} & \shortstack{$0.9172\pm0.0402$} & \shortstack{$0.8394\pm0.0521$} & \shortstack{$0.8381\pm0.0525$} & \shortstack{$0.8006\pm0.0491$}\\  [1.2ex]  
\shortstack{Max Softmax} & \shortstack{$0.9304\pm0.0326$} & \shortstack{$0.9307\pm0.0388$} & \shortstack{$0.9227\pm0.0382$} & \shortstack{$0.9116\pm0.0427$} & \shortstack{$0.8442\pm0.0504$} & \shortstack{$0.8426\pm0.0516$} & \shortstack{$0.8387\pm0.0510$} \\  [1.2ex]  
\shortstack{Min Entropy} & \shortstack{$0.9312\pm0.0324$} & \shortstack{$0.9328\pm0.0378$} & \shortstack{$\bm{0.9239}\pm0.0376$} & \shortstack{$0.9141\pm0.0415$} & \shortstack{$0.8437\pm0.0506$} & \shortstack{$0.8440\pm0.0517$} & \shortstack{$\bm{0.8415}\pm0.0505$}  \\  [1.2ex] 
\shortstack{Min Energy} & \shortstack{$0.9327\pm0.0341$} & \shortstack{$0.9329\pm0.0337$} & \shortstack{$0.9162\pm0.0384$} & \shortstack{$\bm{0.9206}\pm0.0408$} & \shortstack{$0.8449\pm0.0517$} & \shortstack{$0.8415\pm0.0516$} &
\shortstack{$0.8399\pm0.0508$} \\
    \bottomrule
\end{tabularx}}}

\subfloat[Multimodal Classification\label{tab:output-ensemble-multimodal}]{
\setlength{\tabcolsep}{2pt}
\resizebox{\textwidth}{!}{%
\begin{tabularx}{\textwidth}{c|YYY|YYYY}
    \toprule
    \multirow{2}{*}{Method} & \multicolumn{3}{c|}{IMDb} & \multicolumn{4}{c|}{UCI-HAR}  \\
    \cline{2-8}
     & Anchored-15 & Balanced & Disjoint & Mild & Moderate & Skewed & Disjoint  \\
    \midrule
    \shortstack{Random Selection} & \shortstack{$0.3866\pm0.0042$} & \shortstack{$0.3918\pm0.0083$} & \shortstack{$0.2100\pm0.0136$} & \shortstack{$0.6943\pm0.0161$} & \shortstack{$0.5021\pm0.0082$} & \shortstack{$0.3942\pm0.0074$} & \shortstack{$0.2830\pm0.0027$} \\ [1.2ex]
\shortstack{Soft Voting} & \shortstack{$\bm{0.4225}\pm0.0044$} & \shortstack{$0.4218\pm0.0139$} & \shortstack{$\bm{0.3630}\pm0.0318$} & \shortstack{$\bm{0.7740}\pm0.0238$} & \shortstack{$\bm{0.7248}\pm0.0261$} & \shortstack{$0.4231\pm0.0075$} & \shortstack{$0.4174\pm0.0246$}\\  [1.2ex]  
\shortstack{Logits Averaging} & \shortstack{$0.4196\pm0.0047$} & \shortstack{$\bm{0.4285}\pm0.0106$} & \shortstack{$0.3533\pm0.0348$} & \shortstack{$0.7732\pm0.0123$} & \shortstack{$0.7076\pm0.0293$} & \shortstack{$0.4307\pm0.0220$} & \shortstack{$0.3670\pm0.0388$}\\  [1.2ex]  
\shortstack{Max Softmax} & \shortstack{$0.4196\pm0.0044$} & \shortstack{$0.4103\pm0.0150$} & \shortstack{$\bm{0.3630}\pm0.0318$} & \shortstack{$0.7168\pm0.0098$} & \shortstack{$0.5952\pm0.0417$} & \shortstack{$\bm{0.4528}\pm0.0051$} & \shortstack{$0.4204\pm0.0221$} \\  [1.2ex]  
\shortstack{Min Entropy} & \shortstack{$0.4188\pm0.0043$} & \shortstack{$0.4077\pm0.0142$} & \shortstack{$0.3556\pm0.0373$} & \shortstack{$0.7177\pm0.0115$} & \shortstack{$0.5877\pm0.0410$} & \shortstack{$0.4497\pm0.0066$} & \shortstack{$\bm{0.4211}\pm0.0217$}  \\  [1.2ex] 
\shortstack{Min Energy} & \shortstack{$0.4216\pm0.0040$} & \shortstack{$0.3996\pm0.0135$} & \shortstack{$0.3626\pm0.0308$} & \shortstack{$0.7280\pm0.0208$} & \shortstack{$0.5079\pm0.0481$} & \shortstack{$0.3686\pm0.0253$} & 
\shortstack{$0.3812\pm0.0341$}\\
    \bottomrule
\end{tabularx}}}
\label{tab:output-ensemble}
\end{table*}
\subsection{Comparison of Output Ensemble Methods}
\label{subsec:ensemble-methods}

Table~\ref{tab:output-ensemble} shows the inference accuracy of CE-FI with different output ensemble methods across datasets under the 3-device setting, for each dataset and non-IID condition. In this subsection, we compare results across modalities—including image classification, text, and time-series tasks—to analyze the impact of output aggregation strategies on CE-FI performance. 

For image classification tasks, the effectiveness of ensemble strategies varies depending on the severity of label mismatch. Under relatively mild non-IID settings such as Mild and Moderate, soft voting often achieves the highest accuracy. In these settings, many devices produce reasonably reliable predictions, and averaging their outputs helps reduce individual errors.  
In contrast, under more severe non-IID conditions such as Skewed and Disjoint, selection-based approaches tend to become more competitive. In particular, the Minimum Energy method most frequently achieves the highest accuracy, followed by the entropy-based selection strategy. In such cases, many devices produce unreliable logits for unseen classes, and selecting a single confident prediction helps avoid combining noisy outputs. These observations suggest that selection-based strategies become advantageous as the degree of label mismatch increases.

Across modalities, however, the observed trends show additional differences. Although we observe a similar tendency—aggregation-based approaches often perform well under milder non-IID settings, and selection-based approaches become more effective under stronger non-IID conditions—the impact of the ensemble strategy varies depending on the modality. Compared with image classification tasks, the performance gap between ensemble methods is more pronounced in other modalities, particularly on the UCI-HAR dataset. Notably, the Minimum Energy method, which achieves strong performance in several image classification settings, does not consistently provide competitive results for other modalities, particularly for the time-series task. In a few configurations, certain ensemble strategies may even perform slightly worse than Solo Inference.
These findings suggest that the optimal output ensemble strategy strongly depends on the modality and intrinsic data characteristics, and that applying a single ensemble method uniformly across tasks is suboptimal.

Overall, the results demonstrate that the output ensemble strategy is a critical factor in determining the performance of CE-FI. At the same time, they highlight the necessity of task- and modality-aware design. Furthermore, while appropriate ensemble strategies allow CE-FI to outperform Solo Inference, a gap remains compared to the idealized Proposed w/ Oracle Ensemble setting. This suggests that more suitable confidence estimation mechanisms and potentially learning-based ensemble strategies tailored to the characteristics of CE-FI represent important directions for future work.

\subsection{Dirichlet-Based Partitioning of Pre-training Data}
\begin{figure}[htbp]
  \centering
  
  \subfloat[CIFAR10,  $\alpha=0.5$\label{dirichlet-cifar10-05}]{
  \centering
\includegraphics[width=0.45\linewidth]{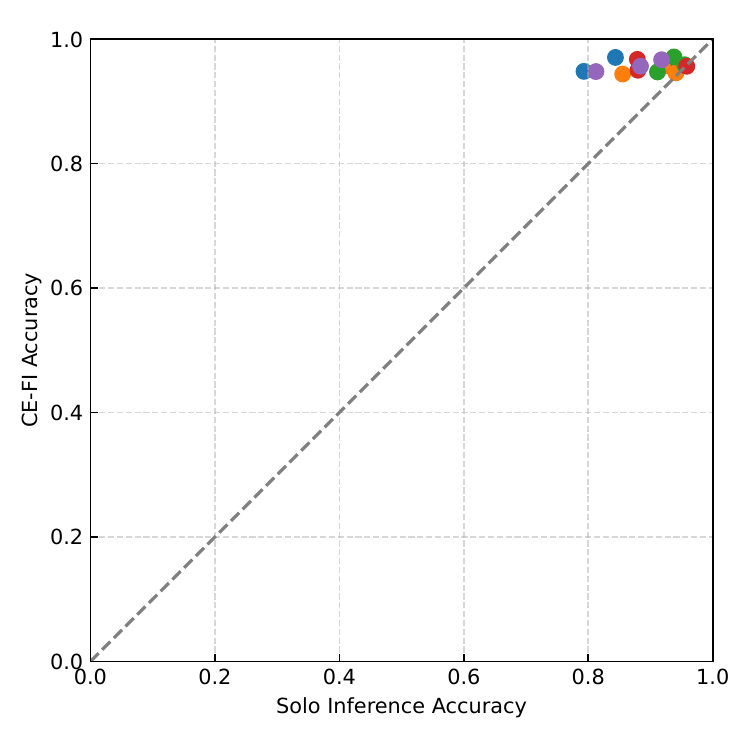}
  }
  \subfloat[CIFAR-10, $\alpha=0.1$\label{fig:dirichlet-cifar10-05}]{
  \centering
\includegraphics[width=0.45\linewidth]{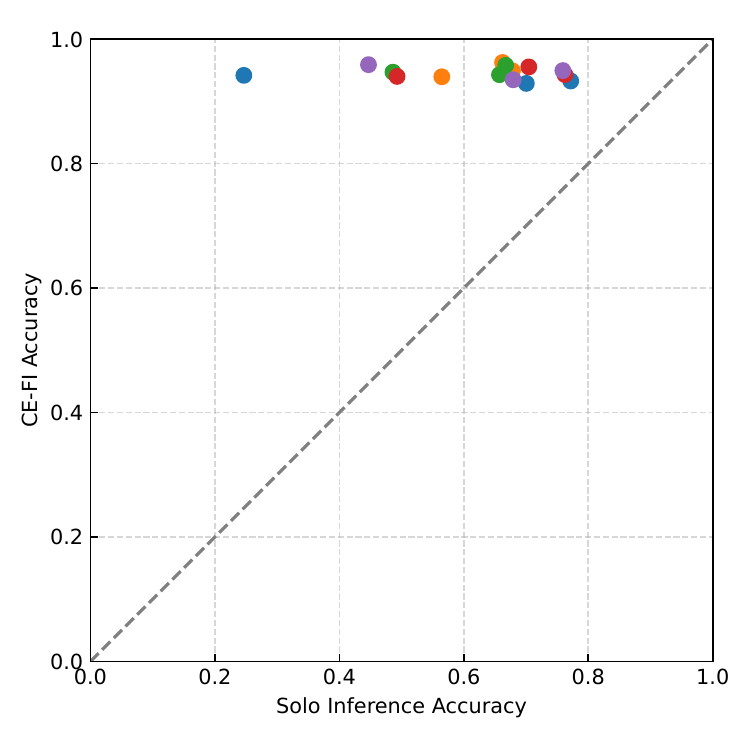}
  }
  

  \subfloat[CIFAR-100-coarse, $\alpha=0.5$\label{dirichlet-cifar100-coarse-05}]{
  \centering
\includegraphics[width=0.45\linewidth]{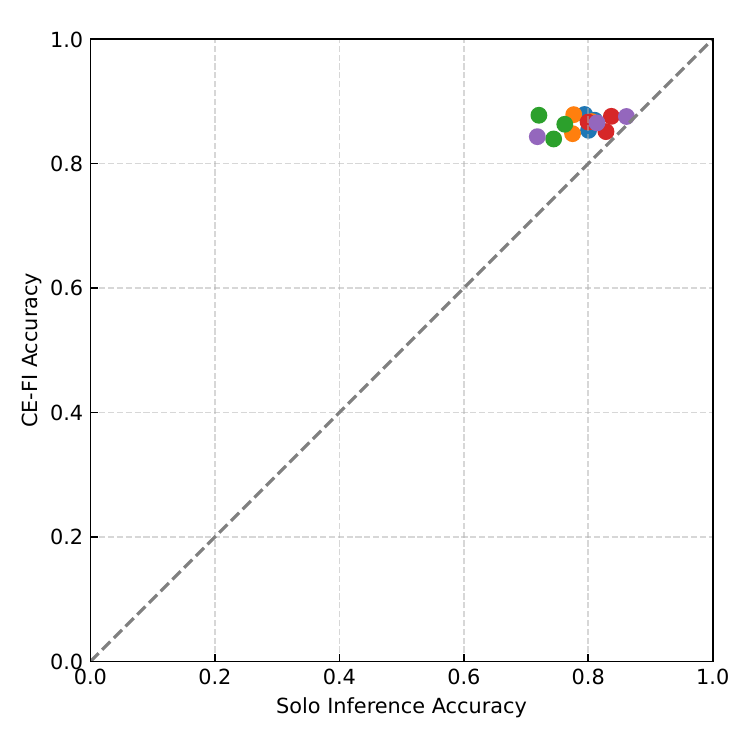}
  }
  \subfloat[CIFAR-100-coarse, $\alpha=0.1$\label{fig:dirichlet-cifar100-coarse-01}]{
  \centering
\includegraphics[width=0.45\linewidth]{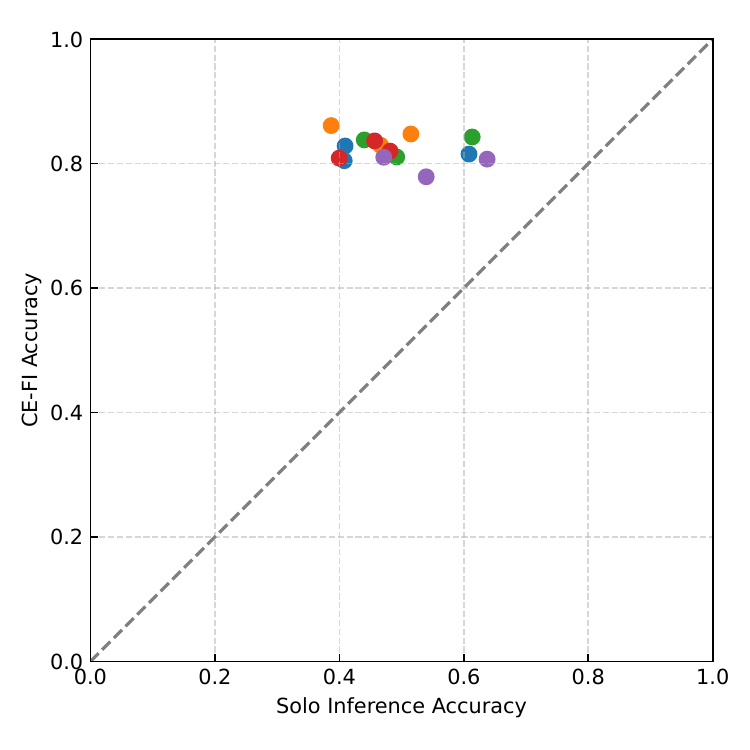}
  }

  \subfloat[CIFAR-100-fine, $\alpha=0.5$\label{dirichlet-cifar100-fine-05}]{
  \centering
\includegraphics[width=0.45\linewidth]{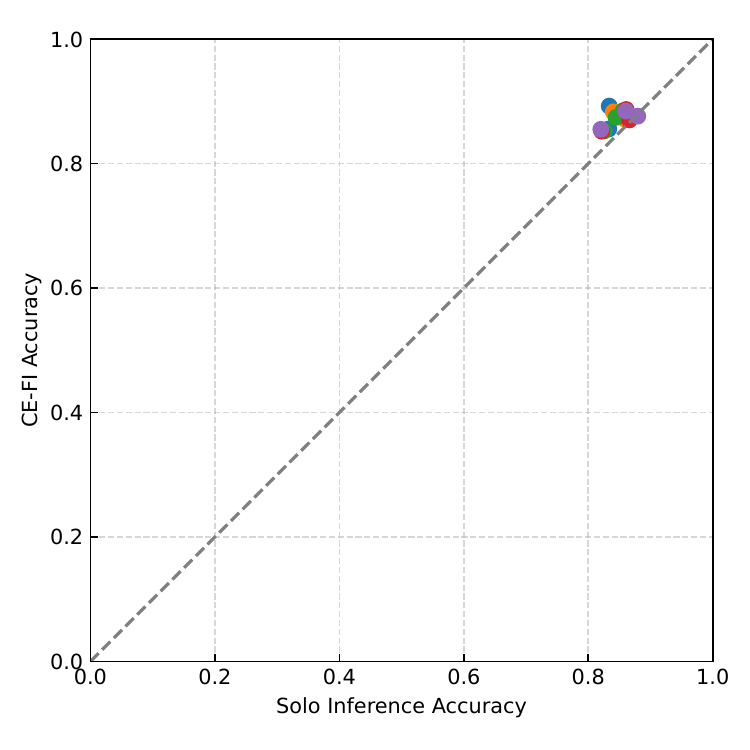}
  }
  \subfloat[CIFAR-100-fine, $\alpha=0.1$\label{fig:dirichlet-cifar100-fine-01}]{
  \centering
\includegraphics[width=0.45\linewidth]{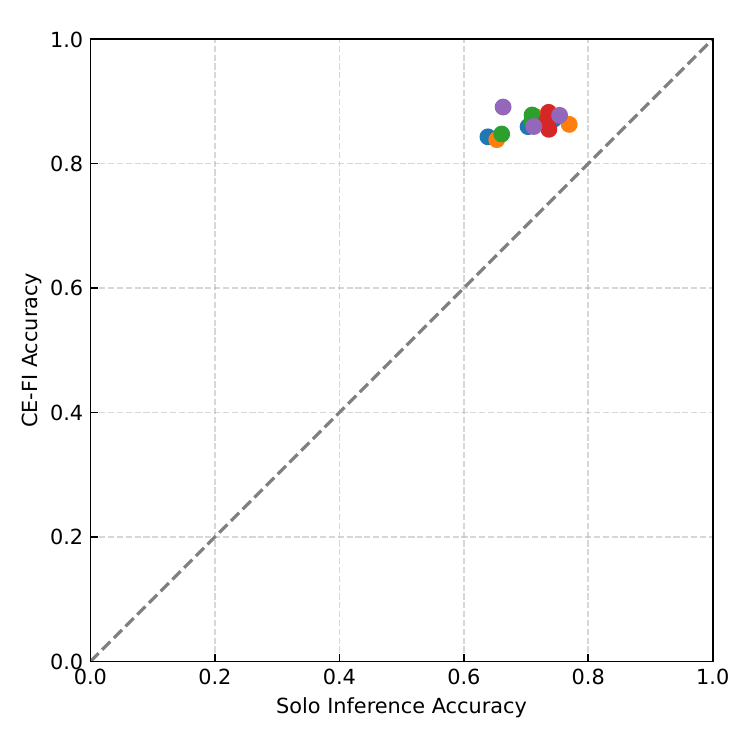}
  }

  \caption{Comparison of accuracy between Solo Inference and CE-FI under Dirichlet partitioning ($\alpha = 0.1, 0.5$), using pretrained encoder (5 trials). Horizontal axis: accuracy of Solo Inference using a device's pretrained model; vertical axis: accuracy of CE-FI using the device's consensus embedding.}
  \label{fig:dirichlet}
\end{figure}
\renewcommand{\arraystretch}{1.2}
\begin{table}[htbp]
\centering
\caption{Difference in classification accuracy between Solo Inference and CE-FI under Dirichlet partitioning (mean $\pm$ std in 5 trials).}
\label{tab:dirichlet}
\begin{tabularx}{\columnwidth}{|c|Y|Y|}
    \toprule
    Dataset & $\alpha=0.1$ & $\alpha=0.5$ \\
    \midrule
    CIFAR-10 & $+0.3270\pm0.1414$ & $+0.0615\pm0.0480$  \\
    CIFAR-100-fine & $+0.1560\pm0.0339$ & $+0.0264\pm0.0143$  \\
    CIFAR-100-coarse & $+0.3343\pm0.0844$ & $+0.0739\pm0.0368$  \\
    \bottomrule
\end{tabularx}
\end{table}

Table~\ref{tab:dirichlet} and Fig.~\ref{fig:dirichlet} summarize the average accuracy gains of CE-FI over Solo Inference and the per-device accuracy scatter plots, where the horizontal axis shows Solo Inference accuracy and the vertical axis shows CE-FI accuracy. We compute the reported average gains by averaging the accuracies of all devices over five trials, each with different data splits sampled from the Dirichlet distribution. The scatter plots in Fig.~\ref{fig:dirichlet} include all device accuracies across the five trials.

Experimental results show that, across all datasets and Dirichlet settings, CE-FI consistently outperforms Solo Inference. In particular, under the strongly non-IID condition corresponding to $\alpha = 0.1$ in the CIFAR-100-coarse setting, CE-FI achieves an average accuracy improvement of approximately 0.33. This indicates that the benefit of cooperative inference becomes especially pronounced when label distributions are highly skewed across devices.
In contrast, under the milder non-IID condition of $\alpha = 0.5$, CE-FI still improves performance across all datasets, but the magnitude of improvement is smaller than in the $\alpha = 0.1$ case. As shown in Fig.~\ref{fig:dirichlet}, this is because device accuracies under Solo Inference are already relatively high when $\alpha = 0.5$, leaving less room for cooperative gains. Importantly, Fig.~\ref{fig:dirichlet} demonstrates that in all five trials, every device achieves higher accuracy with CE-FI than with Solo Inference, confirming that CE-FI provides consistent improvements. Notably, devices with lower Solo Inference accuracy tend to benefit more, suggesting that cooperative inference effectively compensates for performance disparities across devices.

These results demonstrate that CE-FI remains effective not only under manually designed non-IID conditions but also under diverse label distributions generated probabilistically via Dirichlet sampling. Even when substantial imbalance exists in data volume and label coverage across devices, CE-FI mitigates the limitations of individual models through cooperation, highlighting its robustness and effectiveness in a wide range of non-IID distributed environments.

\subsection{Scratch-Trained Models}
\label{sec:scratch}

In the previous evaluations, we primarily demonstrated the effectiveness of CE-FI in settings with pretrained encoders. 
To isolate the impact of pretrained representations, we now consider a more challenging setting in which each device trains its model from scratch using only locally available data. 
This experiment serves as an ablation study to examine how strongly CE-FI depends on the semantic quality of local feature representations.
By removing the benefit of pretrained encoders, we aim to clarify (i) whether CE-FI remains effective under weaker representations and (ii) what factors limit its performance in such scenarios.

\subsubsection{Baseline Comparison}
\begin{figure}[t]
  \centering

  \subfloat[MNIST\label{fig:baseline-mnist}]{
    \centering
\includegraphics[width=\linewidth]{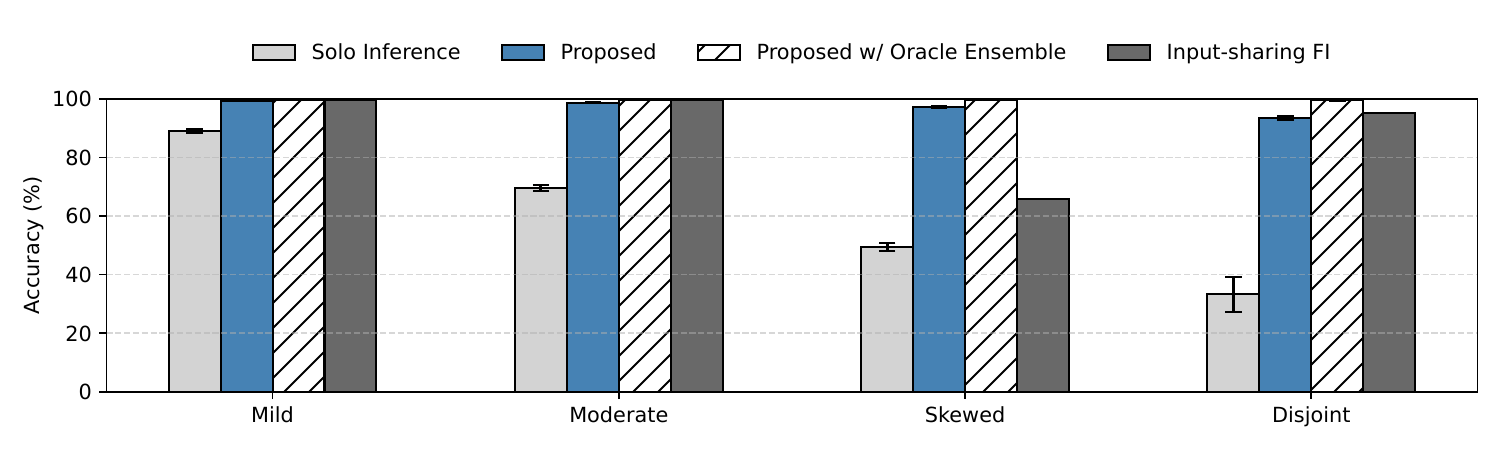}
  }
  \hfill
  \subfloat[FashionMNIST\label{fig:baseline-fmnist}]{
    \centering
\includegraphics[width=\linewidth]{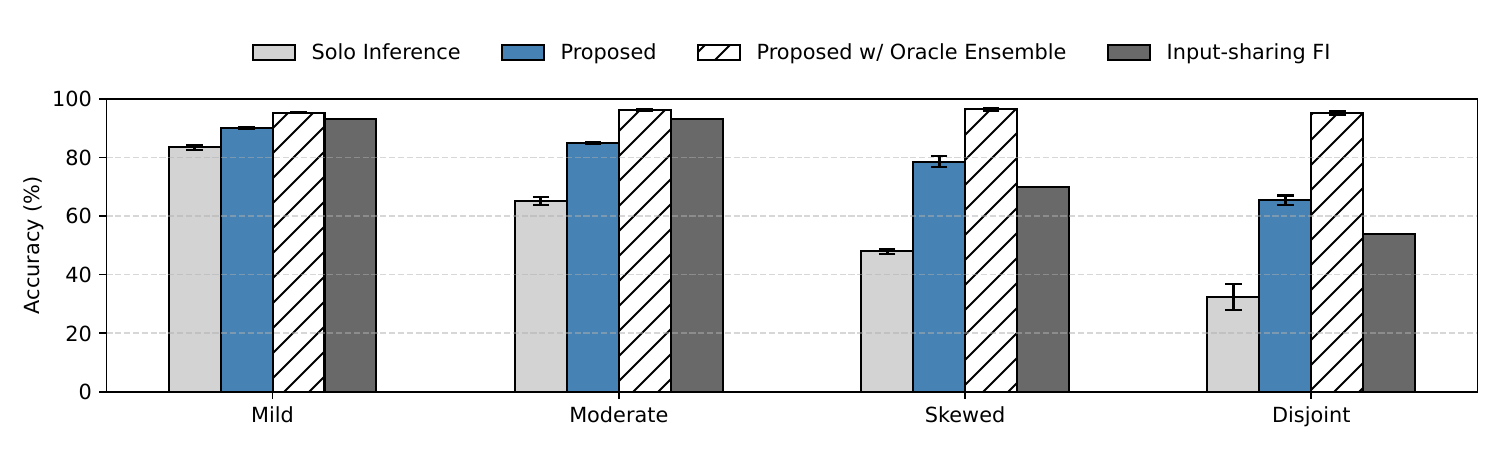}
  }
  \hfill
  \subfloat[CIFAR-10\label{fig:baseline-cifar10}]{
    \centering
    \includegraphics[width=\linewidth]{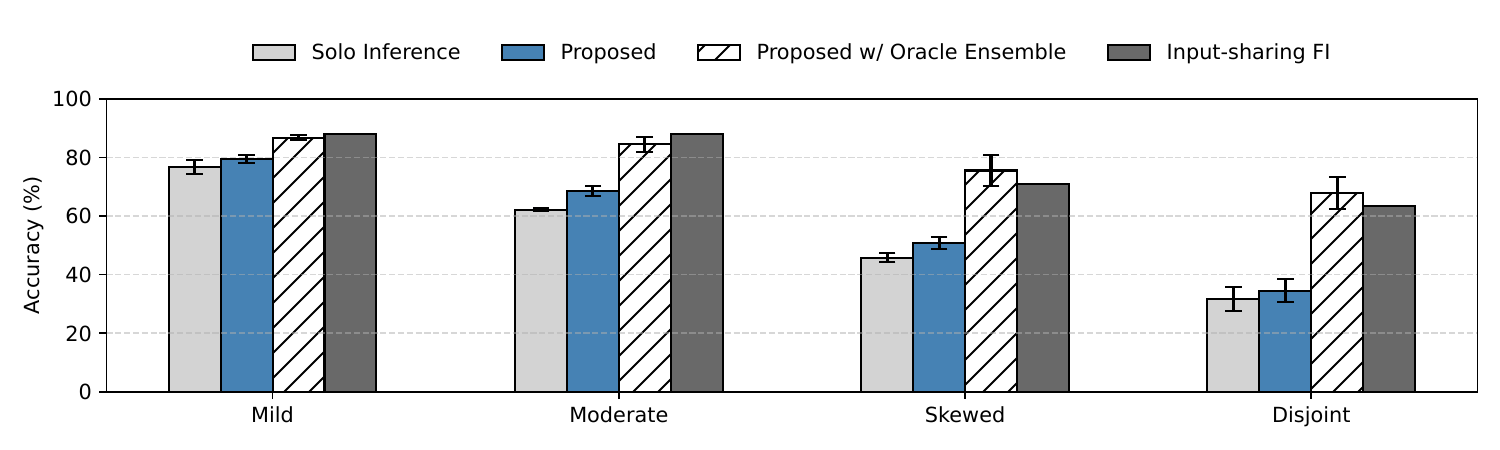}
  }
  
  \caption{Accuracy under various label partition settings with 3 devices: (a) MNIST, (b) FashionMNIST, (c) CIFAR-10 dataset.}
  \label{fig:baseline-comparison-scratch}
\end{figure}

Fig.~\ref{fig:baseline-comparison-scratch} shows the inference accuracy for MNIST, Fashion MNIST, and CIFAR-10 under varying non-IID conditions with three devices.  
As in the pretrained encoder setting, CE-FI consistently outperformed Solo Inference across all datasets and partition conditions, confirming that cooperative inference remains beneficial even without pretrained encoders.  
For relatively simple datasets such as MNIST and Fashion MNIST, CE-FI achieved substantial gains,  even under the Disjoint condition where no labels are shared across devices.  
In contrast, while CE-FI improves accuracy over Solo Inference for CIFAR-10, the performance gain is noticeably smaller than in the pretrained encoder setting, and there remains a large gap compared to the Proposed w/ Oracle Ensemble baseline. 

These findings suggest that while CE-FI remains effective in the scratch setting, its performance gains become increasingly limited as the dataset complexity increases. 
We further investigate the underlying factors contributing to this performance discrepancy in a later analysis of the embedding spaces.

\begin{figure}[t]
  \centering

  \subfloat[MNIST, 5device\label{fig:baseline-dev5-mnist}]{
  \centering
\includegraphics[width=0.45\linewidth]{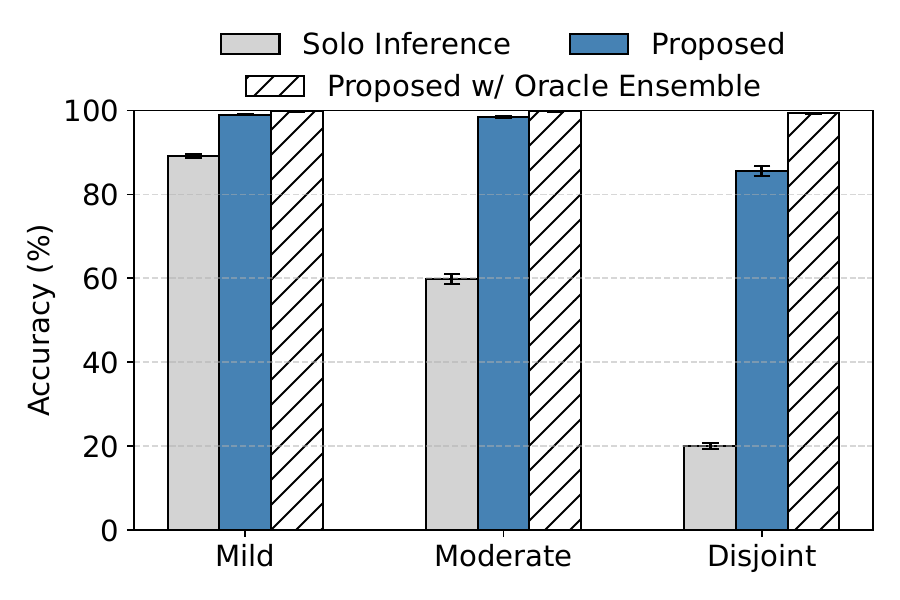}
  }
  \hfill
  \subfloat[MNIST, 10device\label{fig:baseline-dev10-mnist}]{
  \centering
\includegraphics[width=0.45\linewidth]{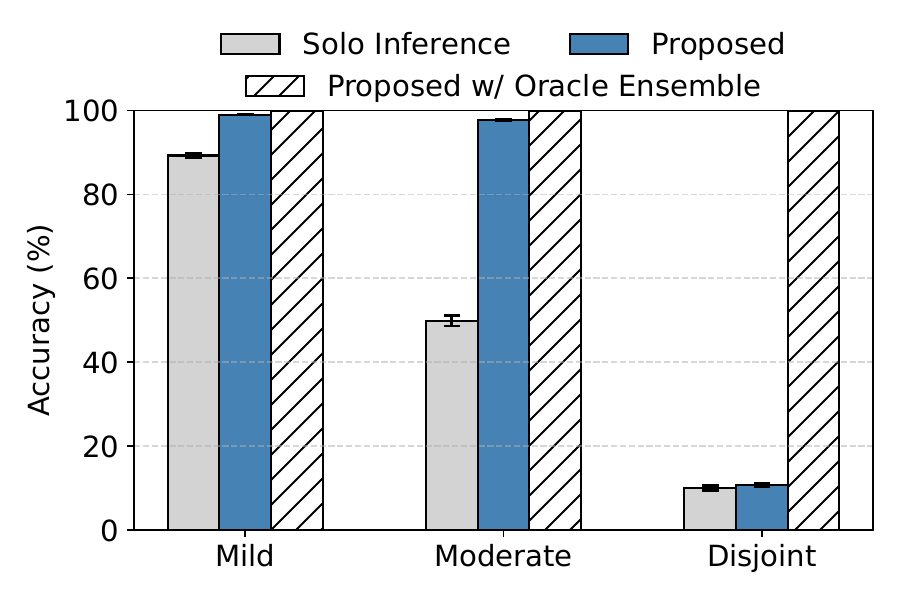}
  }
  \hfill
  \subfloat[Fashion MNIST, 5device\label{fig:baseline-dev5-fmnist}]{
  \centering
    \includegraphics[width=0.45\linewidth]{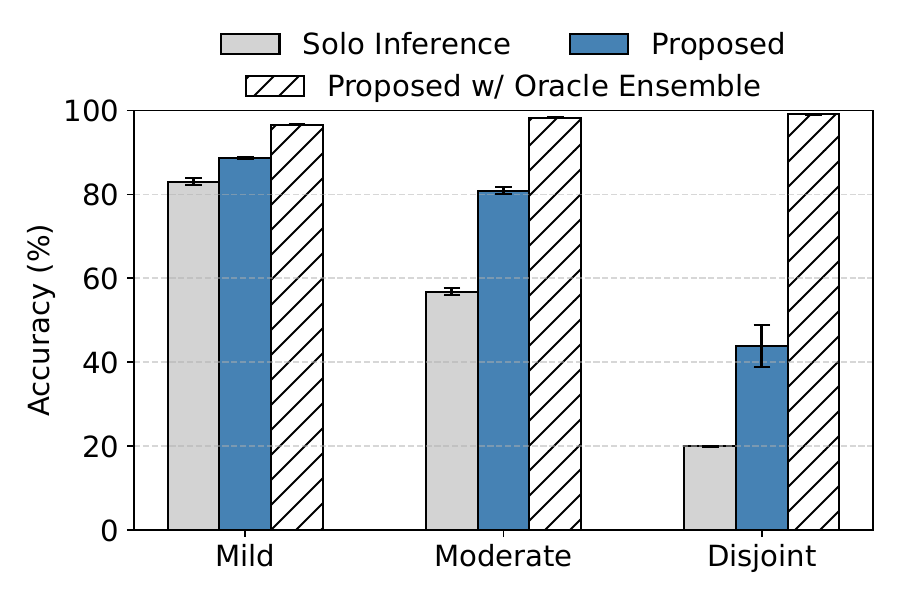}
  }
  \hfill
  \subfloat[Fashion MNIST, 10device\label{fig:baseline-dev10-fmnist}]{
  \centering
\includegraphics[width=0.45\linewidth]{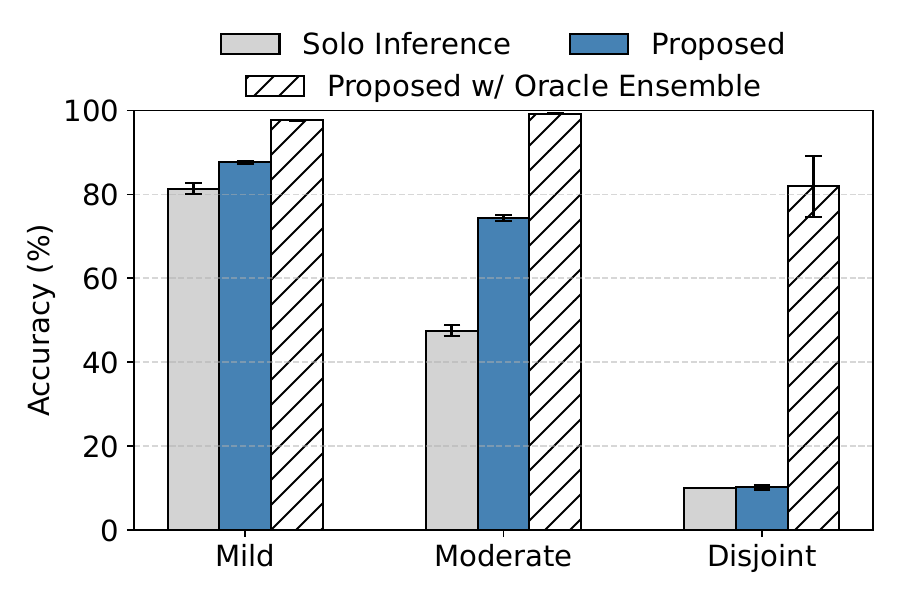}
  }
  \hfill
  \subfloat[CIFAR-10, 5device\label{fig:baseline-dev5-cifar10}]{
  \centering
\includegraphics[width=0.45\linewidth]{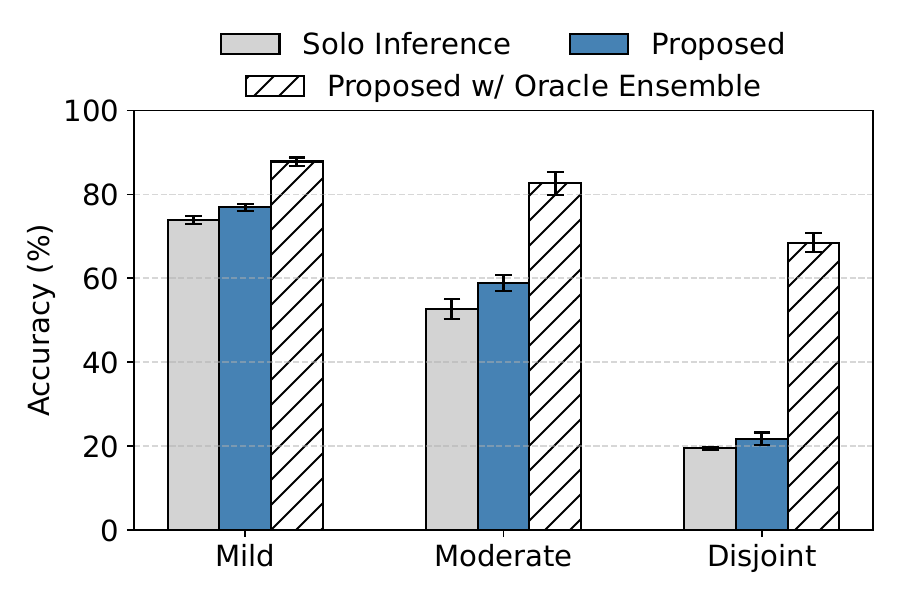}
  }
  \hfill
  \subfloat[CIFAR-10, 10device\label{fig:baseline-dev10-cifar10}]{
  \centering
\includegraphics[width=0.45\linewidth]{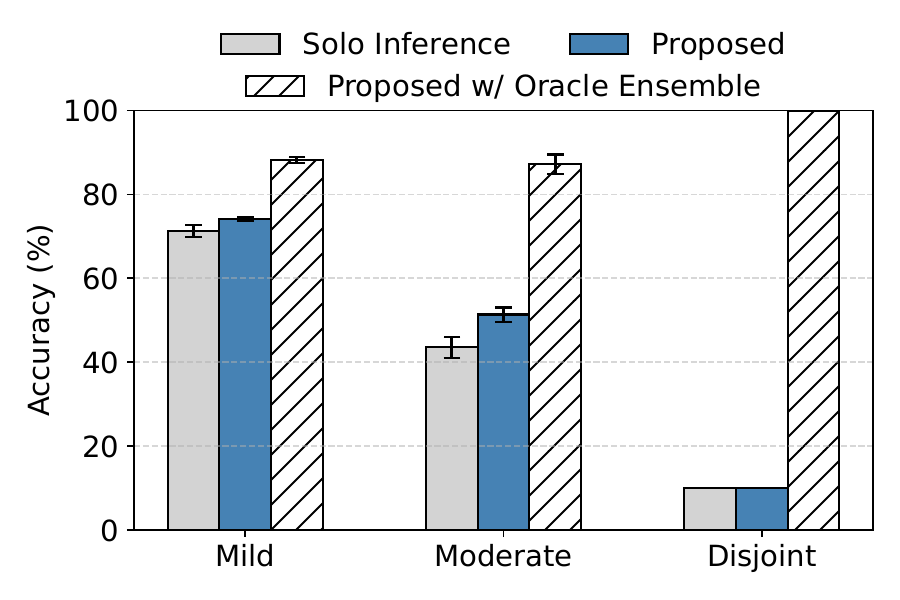}
  }

  \caption{Accuracy under various label partition settings with increasing the number of devices to 5 and 10.}
  \label{fig:baseline-device-expantion}
\end{figure}
\renewcommand{\arraystretch}{1.2}
\begin{table*}[htbp]
\centering
\caption{Difference in classification accuracy between Solo Inference and CE-FI under Dirichlet partitioning (mean $\pm$ std in 5 trials).}
\label{tab:dirichlet_scratch}
\begin{tabularx}{\textwidth}{|c|Y|Y|Y|Y|}
    \toprule
    \multirow{2}{*}{Dataset} & \multicolumn{2}{c|}{$\alpha=0.1$} & \multicolumn{2}{c|}{$\alpha=0.5$} \\
    \cline{2-5}
    & dev 3 & dev 10 & dev 3 & dev 10 \\
    \midrule
    MNIST & $+0.3247\pm0.1371$ & $+0.3598\pm0.1353$ & $+0.0321\pm0.0436$ & $+0.0740\pm0.0624$ \\
    Fashion MNIST & $+0.2294\pm0.0804$ & $+0.1750\pm0.1152$ & $+0.0507\pm0.0471$ & $+0.0791\pm0.0613$ \\
    CIFAR-10 & $+0.1108\pm0.0438$ & $+0.0841\pm0.0346$ & $+0.0718\pm0.0319$ & $+0.1065\pm0.0368$ \\
    \bottomrule
\end{tabularx}
\end{table*}

\subsubsection{Scalability with Number of Devices}

Fig.~\ref{fig:baseline-device-expantion} shows the inference accuracy as the number of devices increases from 3 to 5 and 10.   
While CE-FI maintains performance gains over Solo Inference as the number of devices increases, under the Disjoint condition with 10 devices, CE-FI fails to provide consistent gains. 
In this extreme scenario, some devices are trained on only a single class, resulting in severely limited discriminative capacity at the local level. As a consequence, the cooperative mechanism cannot effectively leverage complementary information across devices, leading to substantial performance degradation.

These observations suggest that while CE-FI scales favorably with the number of devices in typical non-IID scenarios, its effectiveness relies on each device maintaining a minimum level of label diversity.  When label coverage becomes excessively fragmented, cooperative inference becomes substantially restricted.

\subsubsection{Dirichlet-Based Partitioning of Pre-training Data}

\begin{figure}[t]
  \centering
  
  \subfloat[MNIST, 10device, $\alpha=0.1$\label{dirichlet-mnist-10-01}]{
  \centering
\includegraphics[width=0.45\linewidth]{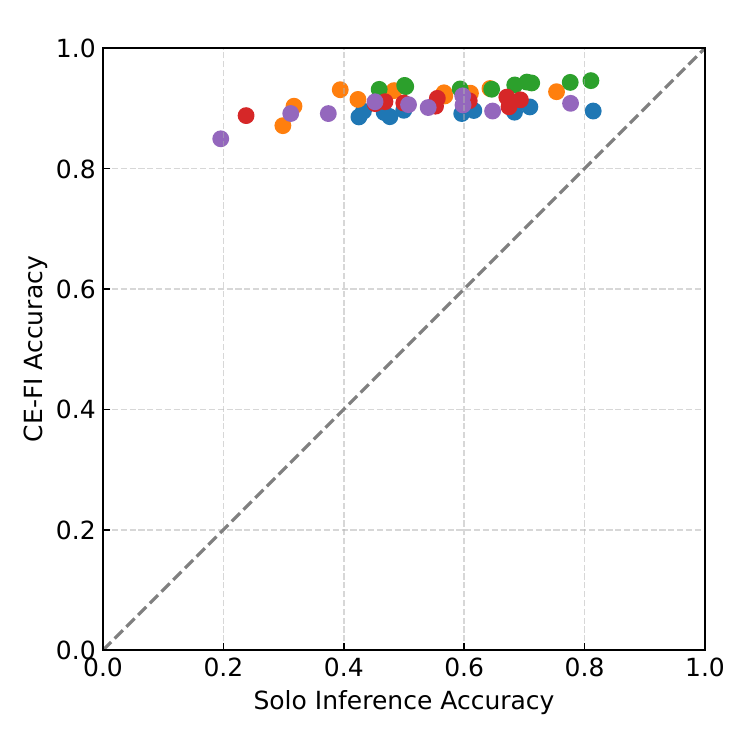}
  }
  \subfloat[MNIST, 10device, $\alpha=0.5$\label{fig:dirichlet-mnist-10-05}]{
  \centering
\includegraphics[width=0.45\linewidth]{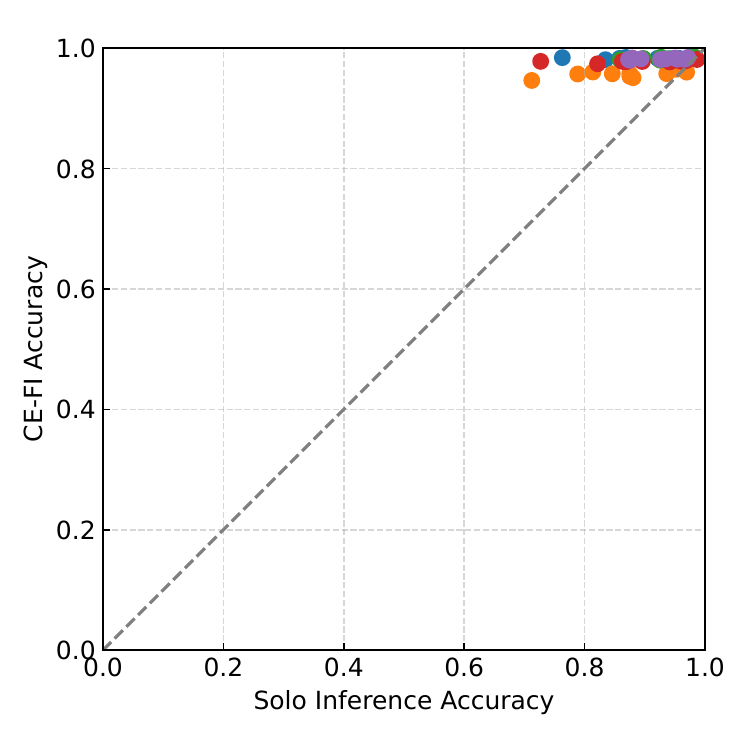}
  }

  \subfloat[CIFAR-10, 10device, $\alpha=0.1$\label{dirichlet-cifar10-10-01}]{
  \centering
\includegraphics[width=0.45\linewidth]{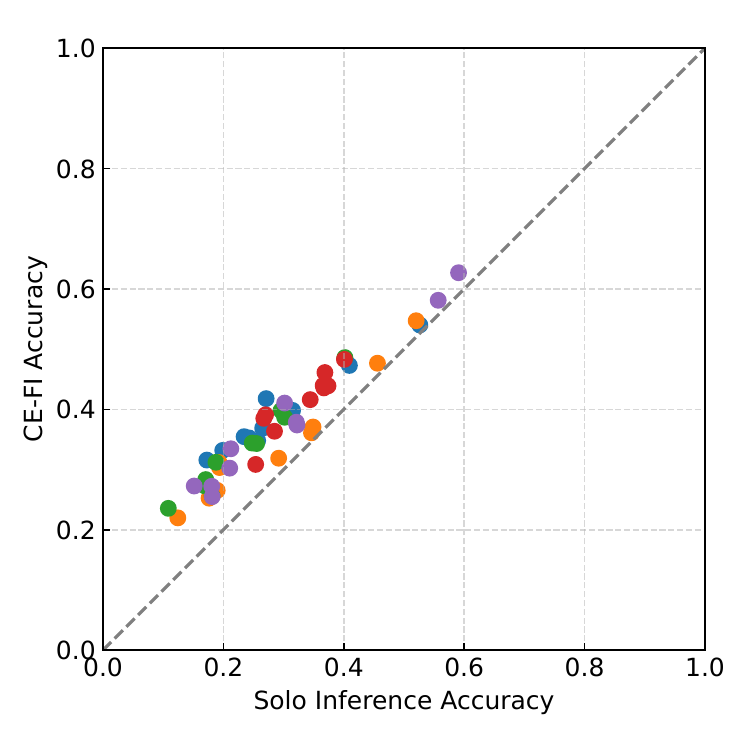}
  }
  \subfloat[CIFAR-10, 10device, $\alpha=0.5$\label{fig:dirichlet-cifar10-10-05}]{
  \centering
\includegraphics[width=0.45\linewidth]{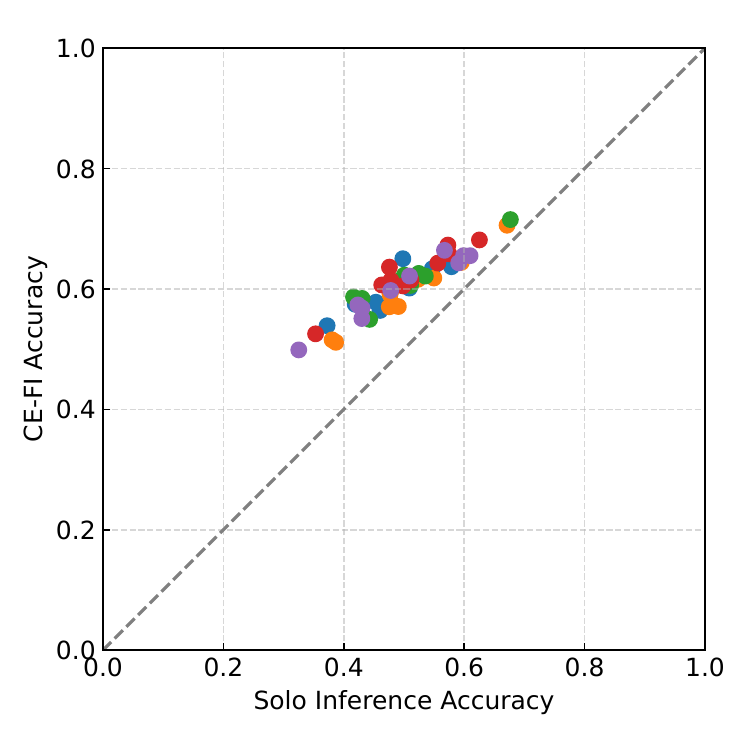}
  }

  \caption{Comparison of accuracy between Solo Inference and CE-FI under Dirichlet partitioning ($\alpha = 0.1, 0.5$) in 5 trials. Horizontal axis: accuracy of Solo Inference using a device's pretrained model; vertical axis: accuracy of CE-FI using the device's consensus embedding.}
  \label{fig:dirichlet_scratch}
\end{figure}

Table~\ref{tab:dirichlet_scratch} and Fig.~\ref{fig:dirichlet_scratch} summarize the average accuracy gains of CE-FI over Solo Inference and the per-device accuracy scatter plots under Dirichlet-based non-IID partitions. We follow the same evaluation protocol as in the pretrained setting, averaging results over five trials with different data splits sampled from the Dirichlet distribution.

Across all datasets and Dirichlet settings, CE-FI consistently outperformed Solo Inference. For instance, on MNIST with $\alpha = 0.1$, CE-FI achieves an average improvement of approximately 0.36. While gains appear smaller for $\alpha = 0.5$, this is largely attributable to the already high Solo Inference accuracy.
However, a more nuanced trend emerges when examining the per-device scatter plots. 
In MNIST, devices with low Solo Inference accuracy are substantially improved under CE-FI, with some gains exceeding 60 percentage points, indicating strong complementary effects across devices. 
In contrast, for CIFAR-10, although CE-FI provides consistent improvements (approximately 10 percentage points on average), the uplift for low-performing devices is more limited. The scatter points tend to align more closely with the diagonal line, suggesting that performance improvements are relatively uniform rather than strongly compensatory.

These observations indicate that, in the scratch setting, CE-FI maintains stable cooperative gains across non-IID partitions, but its ability to substantially compensate for weak local models becomes dependent on dataset complexity.

\subsubsection{Analyzing Consensus Embedding Results}

\begin{figure}[t]
  \centering
  
  \subfloat[MNIST, individual feature\label{fig:dirichlet-mnist-embedding-pretrained}]{
  \centering
    \includegraphics[width=0.45\linewidth]{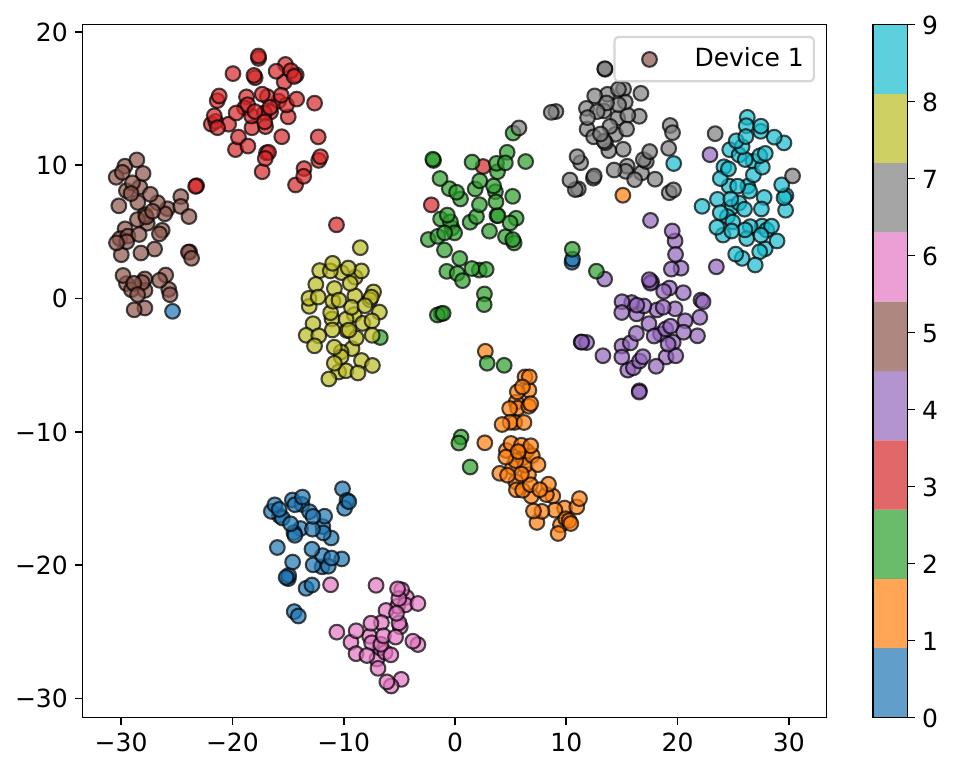}
  }
  \hfill
  \subfloat[MNIST, consensus embedding\label{fig:dirichlet-mnist-embedding-consensus}]{
  \centering
    \includegraphics[width=0.45\linewidth]{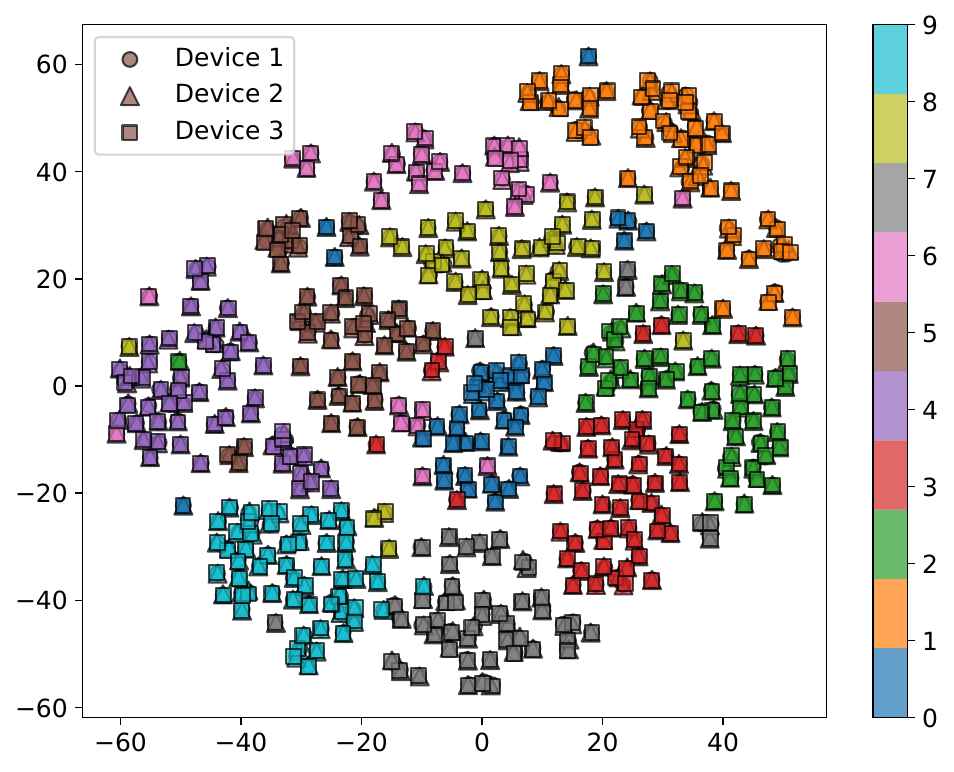}
  }

\hfill
\subfloat[CIFAR-10, individual feature\label{fig:dirichlet-cifar10-embedding-pretrained}]{
    \centering
    \includegraphics[width=0.45\linewidth]{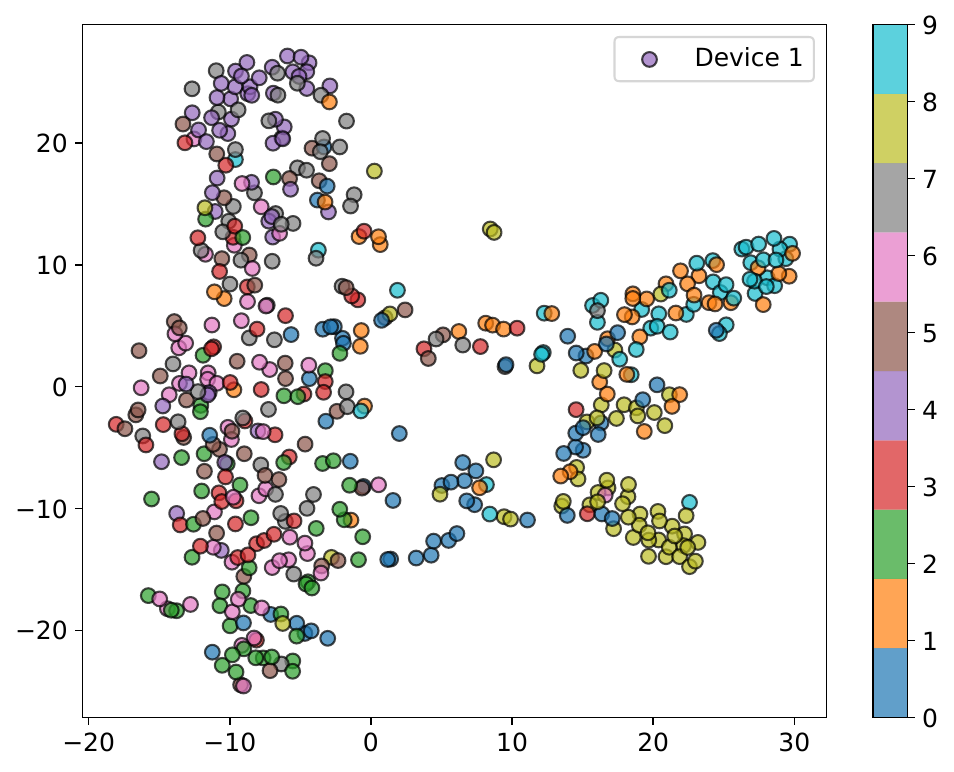}
}
\hfill
\subfloat[CIFAR-10, consensus embedding\label{fig:dirichlet-cifar10-embedding-consensus}]{
    \centering
    \includegraphics[width=0.45\linewidth]{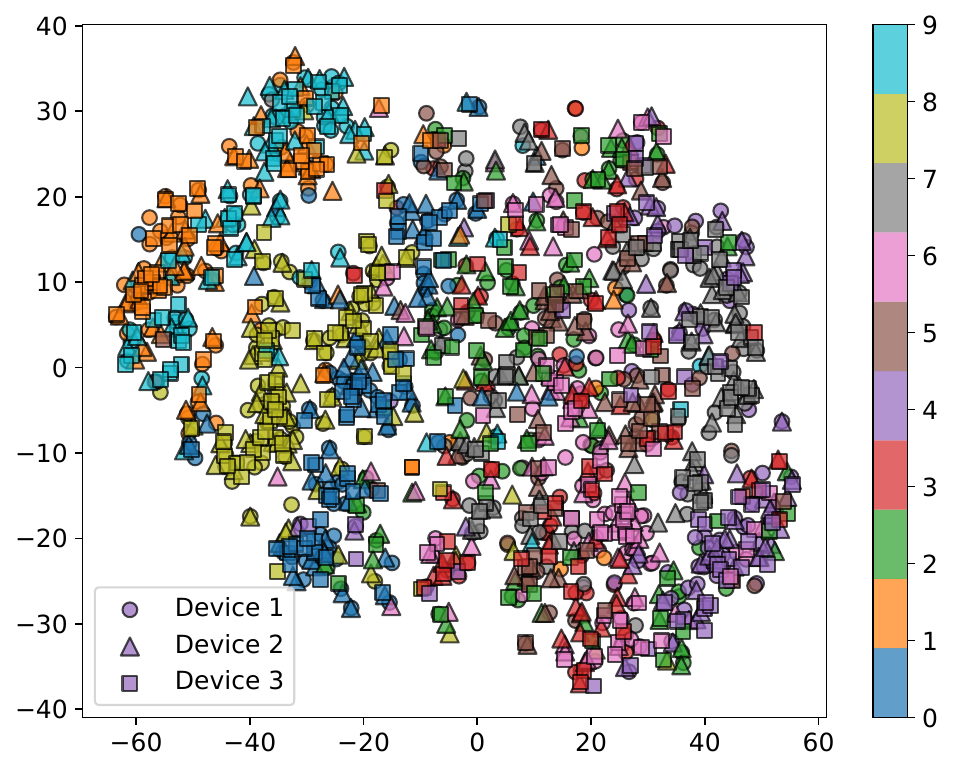}
}
\caption{t-SNE visualization of individual feature spaces from one pretrained model and consensus embedding spaces of pretrained models on MNIST and CIFAR-10 under Dirichlet partitioning ($\alpha=0.1$)  in the 3-device setting.}
\label{fig:dirichlet-embedding}
\end{figure}

To investigate the underlying factors behind the performance differences observed in the scratch setting, we visualized the intermediate feature spaces of individual pretrained models (referred to as individual feature spaces) and the consensus embedding spaces for MNIST and CIFAR-10 using t-SNE dimensionality reduction~\cite{t-SNE}.  
Figs.~\ref{fig:dirichlet-mnist-embedding-pretrained} and~\ref{fig:dirichlet-mnist-embedding-consensus} show the individual feature space of a representative device and the consensus embedding space aggregated from all devices for MNIST, respectively.
In MNIST, we observe that in the individual feature space, well-separated clusters form for each class—even for unseen classes. In the consensus embedding space, class-wise clustering is still clearly observed. Additionally, for each input sample, embeddings from different devices are mapped to nearly identical locations, demonstrating that the CE layer successfully aligns features into a common space. This reflects the intended design of CE-FI: achieving both class-wise structure and inter-device consensus. Notably, this behavior is consistent with the trends observed in the pretrained encoder setting.

Figs.~\ref{fig:dirichlet-cifar10-embedding-pretrained} and \ref{fig:dirichlet-cifar10-embedding-consensus} show the results for CIFAR-10.  
Compared to MNIST, the intermediate feature space of individual models does not exhibit well-formed class-wise clusters; instead, the feature distributions of different classes are substantially intermixed, with only a few classes showing partial separation. 
In the consensus embedding space, inter-device alignment is still achieved to some extent, as the consensus embeddings produced by different devices remain relatively close to one another. However, the class-wise separation is markedly weaker than in MNIST and in the pretrained encoder setting: clusters are less distinct, frequently overlapping, and exhibit noticeable distributional shifts across devices.    
These observations suggest that, in the scratch setting, the CE layer is capable of promoting inter-device alignment, but has a limited ability to construct or reinforce semantically well-separated representations when the underlying local feature spaces lack sufficient structure. 

Taken together, these findings indicate that the effectiveness of CE-FI in the scratch setting depends not only on its ability to align representations across devices, but also critically on the quality of the semantic structure constructed by the consensus embedding space.

\subsection{Domain Shift in Unlabeled Shared Data}
In the previous evaluations, we assumed that the shared unlabeled dataset $D_\mathrm{share}$ used to train the CE and CO layers was drawn from the same domain as the target task. In this subsection, we consider a more challenging setting and analyze the behavior of CE-FI when the shared unlabeled data originate from a different domain.

\begin{figure}[htbp]
  \centering

\includegraphics[width=\linewidth]{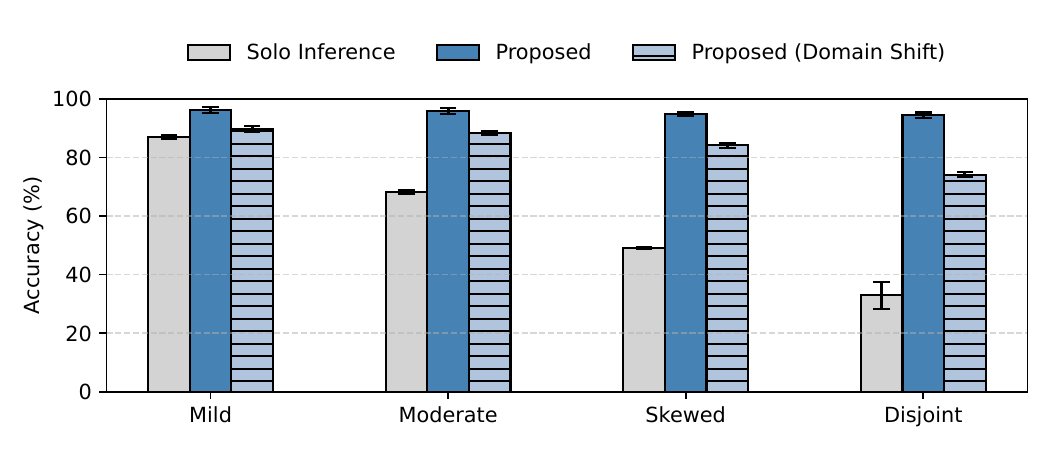}
  
  \caption{Accuracy on CIFAR-10 under domain shift in the shared unlabeled dataset (pretrained encoder, Unified-architecture setting). The CE and CO layers are trained using unlabeled CIFAR-100, while local pretrained models and evaluation are based on CIFAR-10.}
  \label{fig:domain-shift}
\end{figure}

Fig.~\ref{fig:domain-shift} reports the inference accuracy of CE-FI on CIFAR-10 under various label partition settings when CIFAR-100 is used as the domain of the shared unlabeled dataset. In this cross-domain configuration, the pretrained local models and the evaluation task are based on CIFAR-10, whereas the CE and CO layers are trained using unlabeled data from CIFAR-100.
Although CE-FI exhibits moderate accuracy degradation compared to the domain-matched case, CE-FI consistently outperforms Solo Inference across all non-IID settings, even under domain shift.
These findings indicate that while distribution alignment between the shared unlabeled data and the target task is a desirable condition for CE-FI, it is not strictly required. In other words, representation alignment through the CE layer can still operate to some extent even when the shared data are not perfectly task-matched, allowing CE-FI to retain cooperative inference gains.

Overall, this experiment highlights that CE-FI exhibits some dependence on the distribution of the shared unlabeled dataset, but cooperative inference does not collapse entirely under domain mismatch. Since practical deployments may not always provide unlabeled data that perfectly match the target task domain, this experiment supports the robustness and applicability of CE-FI in realistic settings.

\subsection{Bottleneck Analysis}
\label{sec:bottleneck}

In this subsection, we further analyze the factors underlying the performance gaps observed in the previous experiments. Focusing on image classification tasks, we introduce several idealized settings to identify the dominant bottleneck that contributes to the performance gap between CE-FI and conventional FI under certain conditions.

As shown in the analytical results in Sect.~\ref{sec:theoretical-analysis}, the discrepancy between consensus embeddings across devices is a key factor determining whether CE-FI can match the accuracy of Input-sharing FI. The CE layer is trained to align each device's consensus embedding toward the average embedding across devices. In the ideal case, the consensus embedding space would therefore be fully unified by the mean embedding of all devices.
To emulate this ideal condition, we introduce a controlled setting in which we forcibly unify the consensus embeddings used for training the CO layer. Specifically, for each sample $x \in D_\mathrm{share}$, we obtain the consensus embeddings $z_{k,x}$ from all devices and compute their average $\overline{z}_x$. During the CO layer training, this averaged embedding is used in place of the embeddings produced by each device. Furthermore, during inference, we also replace the shared consensus embedding with $\overline{z}_x$, thereby simulating a perfectly unified consensus embedding space across all devices.
In addition, to isolate another potential factor affecting the CO layer, we examine the influence of unknown (OOD) labels contained in the shared unlabeled dataset. Since CE-FI trains the CO layer using unlabeled data, samples belonging to classes unknown to a given device may be included during training. To evaluate whether this contributes to performance degradation, we conduct an idealized experiment in which samples corresponding to device-specific OOD labels are manually excluded from the CO layer training.

\begin{figure}[htbp]
  \centering

\includegraphics[width=\linewidth]{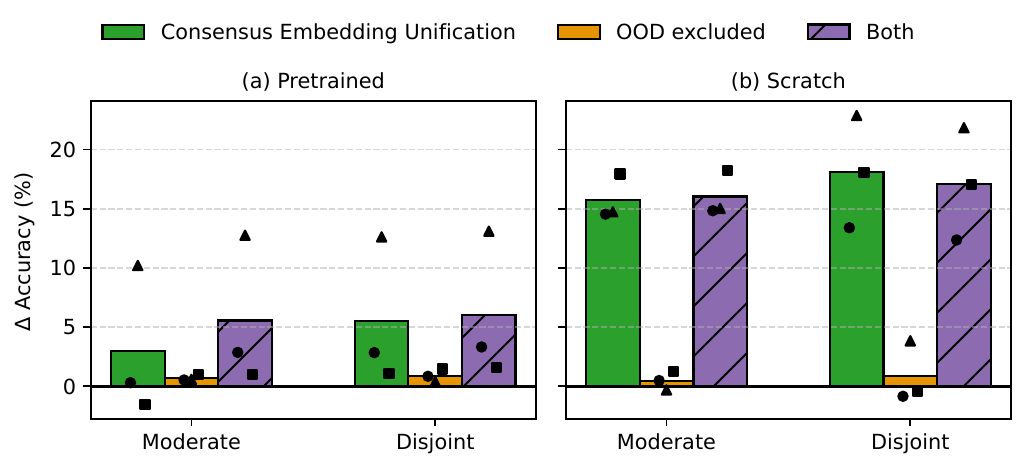}
  
  \caption{Accuracy difference from baseline CE-FI under idealized consensus and OOD-control settings.
Experiments are conducted on CIFAR-100 for the pretrained-encoder (Heterogeneous-architecture) setting and on CIFAR-10 for the scratch setting. Three variants are considered: consensus embedding unification, OOD exclusion during CO training, and their combination. Bars show mean differences, and markers indicate device-wise values.}
  \label{fig:bottleneck-analysis}
\end{figure}

As shown in Fig.~\ref{fig:bottleneck-analysis}, enforcing consensus embedding unification improves inference accuracy over baseline CE-FI in both the scratch and pretrained-encoder settings. The improvement is particularly pronounced in the scratch setting, suggesting that insufficient alignment in the consensus space constitutes a primary performance bottleneck. In contrast, in the pretrained encoder setting, the performance improvement from embedding unification is relatively modest. 
This discrepancy between the two settings provides insight into the underlying bottlenecks. One plausible explanation for this difference is that, in the scratch setting, the head network itself is trained on non-IID local data. Consequently, the types of missing information differ across devices, making it more difficult to construct a well-aligned shared representation space. In contrast, when pretrained encoders are used, the encoders are trained on large-scale and diverse datasets, leading to more naturally aligned representations across devices and facilitating the CE layer learning.

On the other hand, excluding OOD data from the CO layer training does not lead to significant changes in inference accuracy in either setting. The same trend holds even when consensus embeddings are unified. This indicates that the presence of OOD samples in the CO layer training is not the dominant factor affecting performance. Rather, the primary bottleneck of CE-FI lies in representation alignment within the CE layer.
Overall, these results indicate that for CE-FI to approach the performance of an ideal ensemble, improving representation alignment in the CE layer is more critical than strategies such as selective data filtering or OOD exclusion. Therefore, enhancing the quality and semantic separability of the consensus embedding space emerges as the central challenge for further improving CE-FI.

\subsection{Evaluation of Communication Cost and Privacy}
CE-FI enables cooperative inference by exchanging intermediate representations called consensus embeddings, instead of raw inputs or model parameters. 
While this design avoids sharing inputs and model parameters, it introduces communication overhead and potential privacy considerations associated with representation sharing.
In this subsection, we analyze two aspects of this design. First, we quantify the communication cost incurred during both training and inference. Second, we examine the potential privacy risks associated with sharing consensus embeddings through a reconstruction-based analysis.
Unless otherwise specified, we uses the CIFAR-10 dataset with the Moderate label-partitioning setting for the evaluations.

\subsubsection{Training and Inference Costs}
\begin{table}[t]
\caption{Training and Inference Communication Cost of CE-FI.}
\label{tab:training-inference-costs}
\setlength{\tabcolsep}{3pt}

\subfloat[Training cost of CE-FI.]{
\begin{minipage}{0.40\linewidth}
\centering
\begin{tabularx}{\textwidth}{cY}
\toprule
Metric & Value \\
\midrule
Comm. / epoch & $39.8$ MB \\[1.2ex]
Epochs & $45$ \\[1.2ex]
Total comm. & $1.79$ GB \\[1.2ex]
\bottomrule
\end{tabularx}
\end{minipage}
}
\hfill
\subfloat[Inference cost per sample.]{
\begin{minipage}{0.60\linewidth}
\begin{tabularx}{\textwidth}{cYY}
\toprule
Method & Shared Data & Value \\
\midrule
\multirow[c]{2}{*}{\makecell{Input-sharing \\FI}}
& Raw input & $6.2$ KB \\[1.0ex]
& PNG input & $4.6 \pm 0.5$ KB \\

\midrule
CE-FI & \makecell{Consensus \\embedding} & $2.1$ KB \\ [1.0ex] 
\bottomrule
\end{tabularx}
\end{minipage}
}

\end{table}

Table~\ref{tab:training-inference-costs} summarizes the communication
cost of CE-FI in a three-device setting. Training cost is defined as the total communication required to complete one training epoch across all devices, while inference cost corresponds to the total communication required to process a single input sample.  
During training, devices exchange consensus embeddings generated from the shared unlabeled dataset in order to optimize the CE layer.  
One participant temporarily acts as an aggregator: it receives the 256-dimensional embeddings from the other devices, combines them with its own embedding to compute the average embedding used in the contrastive loss, and returns the corresponding embedding-level gradients to the other devices. 
Since only a fixed-size embedding is exchanged for each sample, the communication cost scales linearly with the number of devices. Considering the number of training iterations per epoch, the bidirectional communication, and the batch size of 512, the communication volume per epoch is approximately 39.8~MB. The CE-layer training converges after 45 epochs, resulting in a total communication cost of approximately 1.79~GB.

During inference, the device holding the input computes a consensus embedding and sends it to the other devices. Each receiving device then performs inference using its cooperative output (CO) layer and returns the predicted logits back to the source device. 
For CIFAR-10 with a 256-dimensional consensus embedding and 10 output classes, transmitting the raw input image requires 6.2~KB per sample, while PNG-compressed images require on average 4.6~KB. In contrast, the proposed consensus embedding requires only 2.1~KB per sample. 

Reducing communication costs is not the primary objective of CE-FI; the framework aims to enable cooperative inference without sharing inputs or model parameters. Nevertheless, in the image-classification setting considered here, transmitting a 256-dimensional consensus embedding required less communication per sample than transmitting the input itself. Whether similar communication savings can be achieved in other modalities depends on the input representation and the dimensionality required for the shared embedding, and remains an important direction for future work.

\subsubsection{Privacy Evaluation}
Although CE-FI avoids sharing raw input data and model parameters, devices exchange intermediate representations referred to as consensus embeddings. While such representations generally reduce privacy risks compared with transmitting raw inputs, they may still retain information about the original inputs. Prior work on split learning and split computing has shown that shared intermediate representations can be exploited in data reconstruction attacks (DRAs), where an adversary attempts to reconstruct the original input from transmitted features~\cite{erdougan2022unsplit,pasquini2021unleashing,xu2024stealthy-fora}.
Although various defense mechanisms have been proposed, preventing information leakage from intermediate representations remains an open research problem. Therefore, similar privacy concerns may also arise in inference frameworks that exchange intermediate representations, including Edge Ensemble and CE-FI.

To examine this risk in CE-FI, we conduct a reconstruction attack experiment inspired by prior DRAs.
Using the shared unlabeled dataset, we evaluate whether an adversary can reconstruct the original input from the exchanged consensus embeddings. We evaluate this attack under two training settings: pretrained encoder and scratch settings.

In the attack scenario, the adversary is assumed to have access to the consensus embeddings exchanged among devices and to the shared unlabeled dataset used to train the CE and CO layers. During CE-layer training, embeddings corresponding to samples from the shared dataset are exchanged across devices. As a participant in this protocol, the adversary can obtain pairs of embeddings and their corresponding input images. Using these pairs, the adversary trains a decoder network to reconstruct the original image from the shared representation using pixel-wise mean squared error (MSE).
The decoder architecture follows a DCGAN-inspired design~\cite{radford2015dcgan}. The consensus embedding is first expanded by a linear layer and reshaped into a low-resolution latent feature map and then upsampled using transposed convolution, batch normalization, and ReLU layers. A final transposed convolution with sigmoid activation produces an image with the same resolution as the original input. 
For reference, we additionally evaluate reconstruction from intermediate features. Although intermediate features are not exchanged between devices in the CE-FI protocol, this comparison helps illustrate the relative information content of different representations. The same reconstruction architecture and loss function are used in this setting.

\begin{figure}[htbp]
  \centering
  
  \subfloat[pretrained encoder, intermediate feature\label{fig:recon-vit-intermediate}]{
  \centering
    \includegraphics[width=0.45\linewidth]{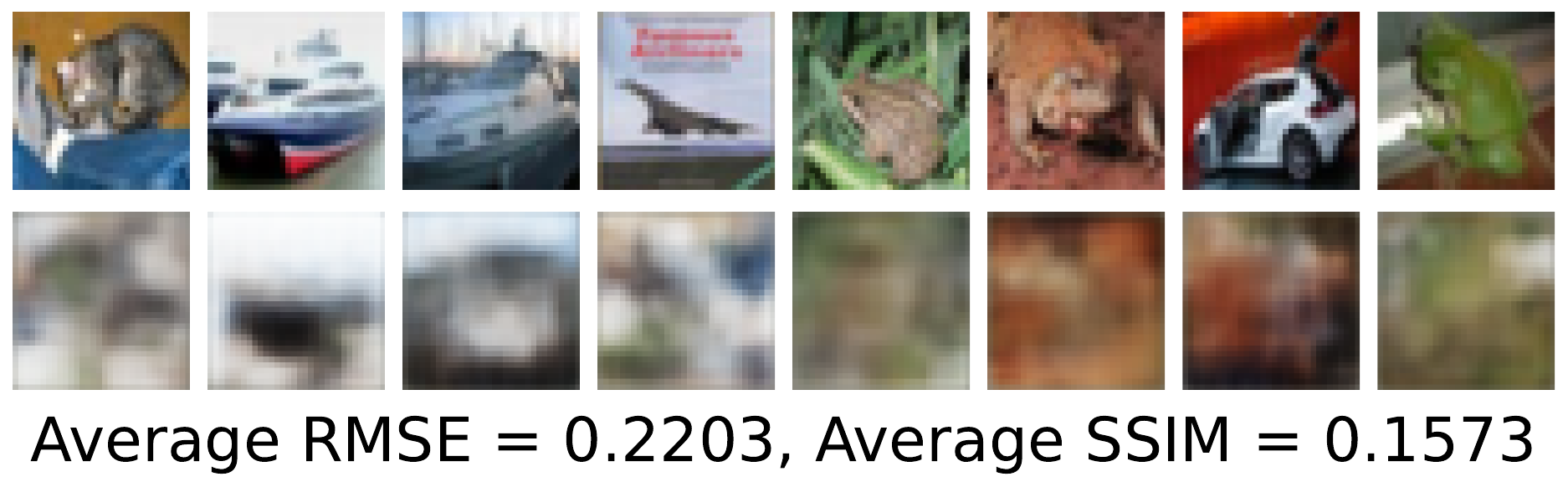}
  }
  \hfill
  \subfloat[pretrained encoder, consensus embedding\label{fig:recon-vit-consensus}]{
  \centering
    \includegraphics[width=0.45\linewidth]{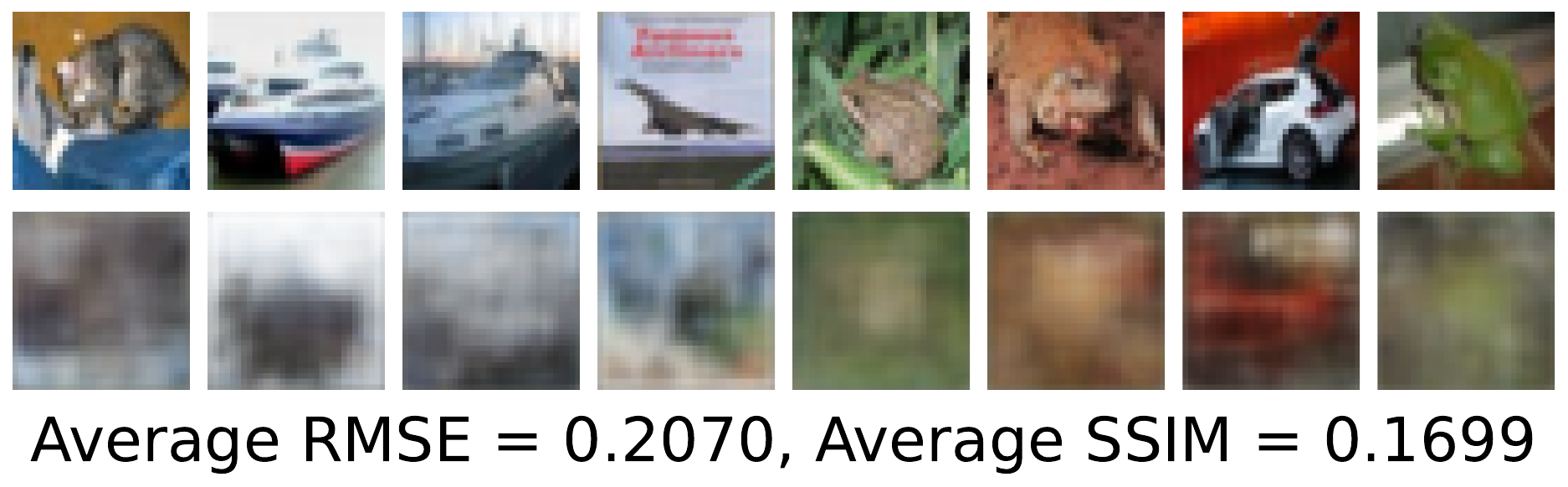}
  }
  \hfill
  \subfloat[scratch encoder, intermediate feature\label{fig:recon-scratch-intermediate}]{
  \centering
    \includegraphics[width=0.45\linewidth]{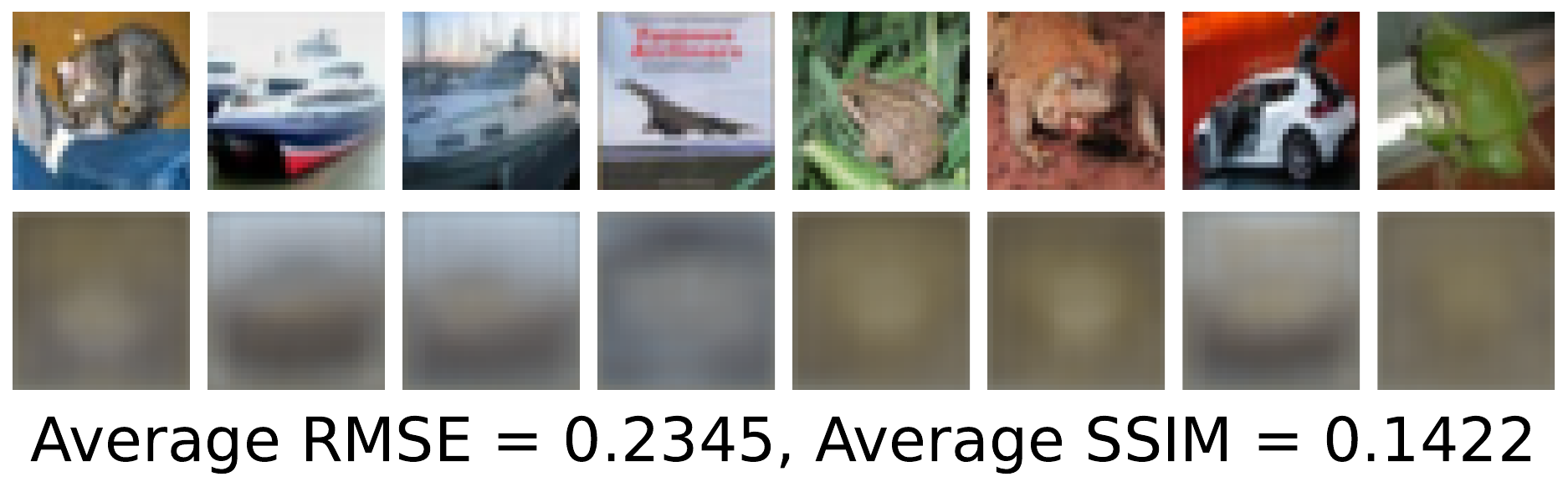}
  }
  \hfill
  \subfloat[scratch encoder, consensus embedding\label{fig:recon-scratch-consensus}]{
  \centering
    \includegraphics[width=0.45\linewidth]{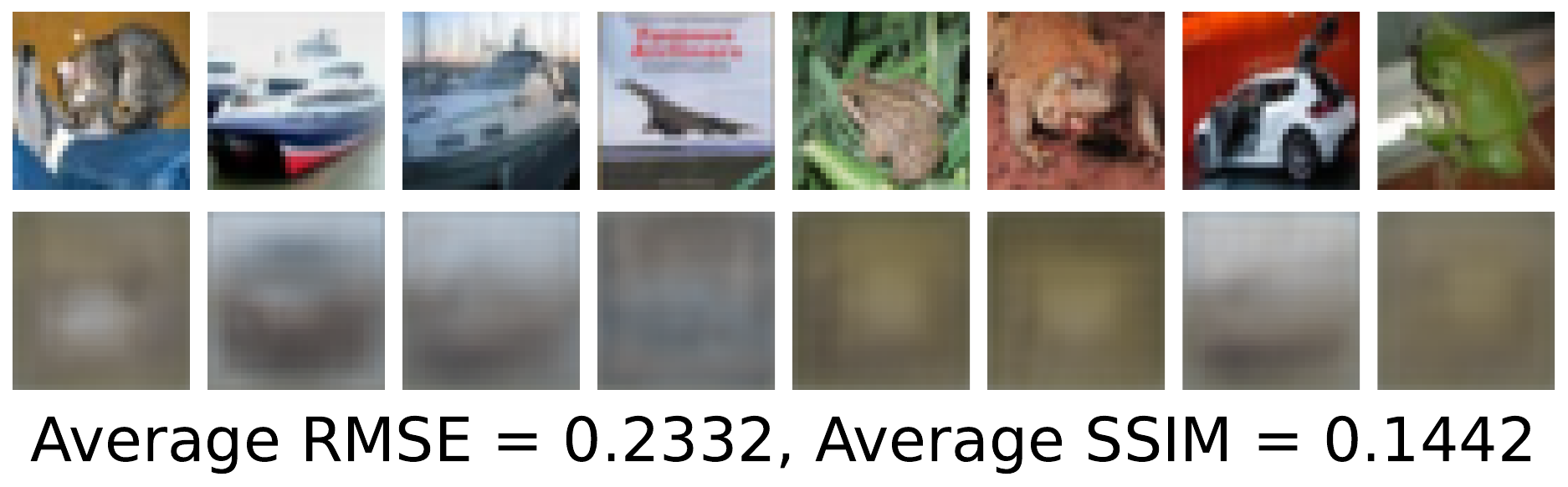}
  }
  
\caption{Reconstruction attack results on CIFAR-10. 
For visualization, eight representative samples randomly selected from the test set are shown.
The top row shows the original images, and the bottom row shows the reconstructed images. 
Results are shown for two representation types—consensus embeddings and intermediate features—and two training settings: pretrained encoder and training from scratch.}
\label{fig:privacy-recon}
\end{figure}

Fig.~\ref{fig:privacy-recon} presents representative reconstruction examples together with the average RMSE and SSIM values computed over the CIFAR-10 test set. 
As shown in the figure, accurately reconstructing the original input from either intermediate features or consensus embeddings is difficult in both settings. 
In particular, when using models trained from scratch, the reconstructed outputs resemble visually meaningless noise. Even in the pretrained encoder setting, no reconstructions were observed that clearly revealed the content or class identity of the original input.
This observation is also supported by quantitative metrics reported in Fig.~\ref{fig:privacy-recon}. 
Across all settings, the reconstruction quality remains low, with RMSE values around 0.20--0.23 and SSIM values below 0.17, indicating limited structural similarity with the original images. 
These findings suggest that, under the present experimental setup, reconstructing the original input from the shared intermediate features or consensus embeddings is not straightforward, indicating that sharing such representations may reduce the direct exposure of original inputs compared with directly sharing raw data. While this experiment provides an empirical indication of the information leakage risk from the exchanged representations, a comprehensive analysis of potential privacy attacks is left for future investigation.

\section{Conclusion}
We proposed CE-FI, a novel framework for cooperative inference that enables multiple pretrained models to perform joint inference without sharing model parameters, raw inputs, or requiring a common encoder. By leveraging unlabeled intermediate representations in a self-supervised manner, CE-FI eliminates the need for labeled data during cooperative training.
Extensive experiments across image, text, and time-series tasks demonstrate that CE-FI consistently outperforms Solo Inference and, in many settings, approaches the performance of conventional FI despite operating under significantly stricter constraints. Additional analyses reveal that representation alignment in the CE layer is the primary factor influencing performance, while reconstruction experiments suggest that sharing intermediate representations does not trivially expose raw inputs under the examined settings.

Future work will focus on improving representation alignment and developing more robust confidence estimation and ensemble strategies to further close the gap to ideal cooperative inference. In addition, several extensions remain for practical deployment. These include understanding the scalability of CE-FI in larger cooperative environments, improving the efficiency of training and inference in terms of communication and computation, enhancing robustness against more advanced privacy attacks, and extending the theoretical analysis beyond the specific architectural assumptions considered in this work.

\bibliographystyle{IEEEtran}
\bibliography{references}

\end{document}